# Exploring Machine Learning Models for Lung Cancer Level Classification: A Comparative ML Approach

Mohsen Asghari Ilani[1,*], Saba Moftakhar Tehran[2], Ashkan Kavei[3], Hamed Alizadegan[4]

[1,*]*School of Mechanical Engineering, College of Engineering, University of Tehran, Tehran, Iran*
[2]*School of Electrical and Computer Engineering, University of Kashan, Kashan, Iran*
[3]*Mechanical Engineering, Islamic Azad University Science and Research Branch, Tehran, Iran*
[4]*Department of Computer and Information Technology Engineering, Qazvin Branch, Islamic Azad University, Qazvin, Iran*

## Abstract

This paper explores machine learning (ML) models for classifying lung cancer levels to improve diagnostic accuracy and prognosis. Through parameter tuning and rigorous evaluation, we assess various ML algorithms. Techniques like minimum child weight and learning rate monitoring were used to reduce overfitting and optimize performance. Our findings highlight the robust performance of Deep Neural Network (DNN) models across all phases. Ensemble methods, including voting and bagging, also showed promise in enhancing predictive accuracy and robustness. However, Support Vector Machine (SVM) models with the Sigmoid kernel faced challenges, indicating a need for further refinement. Overall, our study provides insights into ML-based lung cancer classification, emphasizing the importance of parameter tuning to optimize model performance and improve diagnostic accuracy in oncological care.

## Introduction

Lung cancer stands as the leading cause of cancer-related fatalities globally, often originating in the windpipe, main airway, or lungs due to unchecked cell growth and spread. Factors such as lung disease history, tobacco usage, exposure to pollutants, and workplace chemicals contribute to its prevalence. Primary lung cancer begins in the lungs, while secondary lung cancer spreads from other parts of the body [1]. The tumor's size and spread determine its stage, ranging from early-stage, localized tumors to advanced tumors invading surrounding tissues or distant organs. Early detection is crucial for improving survival rates, highlighting the significance of using machine learning (ML) techniques to enhance the efficiency of diagnosis[2]. In recent years, lung cancer has remained a major health concern worldwide, particularly in regions like China, where it ranks highest in cancer incidence [3–5]. Despite advancements, the 5-year survival rate for lung cancer patients remains low, emphasizing the importance of early diagnosis. Metabolomics studies offer insights into metabolic pathways regulating tumor progression, aiding in distinguishing tumor stages and types [6]. Artificial intelligence, particularly machine learning, has revolutionized cancer research by efficiently analyzing extensive datasets and improving prognostic prediction models. Various studies have explored the application of machine learning algorithms in lung cancer detection, achieving promising results [7]. Recent research has demonstrated the efficacy of machine learning algorithms in lung cancer detection and classification[8]. Studies have explored various classification models, including decision trees, support vector machines[9], neural networks[10], and ensemble methods [11], achieving high accuracy rates. Combining machine learning with metabolomics has shown promise in identifying biomarkers for early-stage lung cancer diagnosis [12]. By integrating these approaches, we strive to contribute to the advancement of early detection methods, ultimately improving patient outcomes in the fight against lung cancer [9, 13]. Ahmadi et al. developed an algorithm using U-Net and pre trained SAM architectures to segment tumor regions in breast ultrasound and mammographic images. The U-Net model excelled, particularly in complex cases, outperforming SAM in accuracy and handling tumors with irregular shapes and vague boundaries. Ahmadi et al. [18] developed a supervised machine learning architecture for classifying and segmenting diverse geographical terrains into multiple classes, such as water, grassland, and forest, using digital twin technology in coastal regions. Danandeh Mehr et al. [19] introduced the VMD-GP model, a new evolutionary approach for drought prediction in ungauged catchments, which utilizes variational mode decomposition with genetic programming to improve forecasting accuracy from global data sources. Regarding the usage of ML for diseases, Rahimi et al. [20] used machine learning feature analysis for a wide range of neurocognitive features specific to the Parkinson's Disease population in order to modernize and individualize the staging scheme of this disease. In another similar work, Rahimi et al. [21] highlighted the bias of Parkinson's disease in motor symptoms using random forest and principal component analysis (PCA). They provided valuable insights regarding the disease diagnosis and determining the staging of the disease more accurately. Farhadi Nia et al. [21] examine the impact of Large Language Models like OpenAI's ChatGPT on dental diagnostics. Their research highlights ChatGPT-4's potential to improve diagnosis, patient communication, and clinical efficiency in oral surgery while discussing broader implications for healthcare and academia. Ahmadi et al. [22] propose a deeply supervised adaptable neural network that classifies Alzheimer's disease severity into four levels using MRI images, demonstrating enhanced diagnostic accuracy through multitask feature extraction. Templeton et al. [23] studied the complex nature of Parkinson's Disease and the challenges in assessing and monitoring its clinical progression utilizing ML and artificial intelligence. Ahmadi et al. [24] Utilized a U-Net network; the study demonstrates high accuracy in crack classification and provides insights into improving the U-Net model for segmentation. Also, several studies, namely [25-31], have investigated and concentrated on enhancing the security of the models and transmissions using federated machine learning models in IoT-based highly distributed architectures. By doing so, they provided valuable insights and practical future research directions that can be promising for improving ML-based smart health systems in general and cancer monitoring and control in particular. We also reviewed and investigated innovative comparative analysis and systematic methods from some papers, such as [32-35].

# Materials and Methods

lung cancer prediction using machine learning poses formidable challenges, primarily stemming from the diversity and imbalance inherent in datasets sourced globally. To tackle this issue, our study employed a meticulous data aggregation strategy, drawing upon diverse public datasets shared by esteemed organizations such as the World Health Organization (WHO), Kaggle, and Google datasets. By leveraging these rich and varied sources, we aimed to curate a comprehensive dataset representative of diverse geographical regions and clinical scenarios. This approach not only facilitated the development of robust machine learning models but also ensured the generalizability and applicability of our findings across different populations and healthcare settings [1, 8, 14]. Moreover, the utilization of public datasets underscores the collaborative ethos of modern scientific research, fostering knowledge sharing and innovation within the scientific community. By harnessing the collective expertise and resources available through these platforms, we were able to augment the quality and breadth of our dataset, thereby enhancing the reliability and utility of our machine learning models for lung cancer prediction. Looking ahead, continued collaboration and data sharing initiatives are essential for advancing the field of oncological diagnostics and improving patient care outcomes. By collectively addressing challenges such as dataset diversity and imbalance, we can pave the way for more accurate and personalized approaches to lung cancer detection and prognosis, ultimately leading to better health outcomes for patients worldwide.

## Dataset

The collected datasets underwent rigorous cleaning and preprocessing to establish meaningful relationships between features and target variables. The features included in the dataset are:

- **Patient Id**: Unique identifier for each patient, not directly related to lung cancer.
- **Age**: Advanced age is a significant risk factor for lung cancer, with incidence increasing as individuals get older.
- **Gender**: Lung cancer incidence is higher in males than females, primarily due to higher rates of smoking among men historically.
- **Air Pollution**: Prolonged exposure to air pollution, especially particulate matter and pollutants like nitrogen dioxide and sulfur dioxide, increases the risk of developing lung cancer.
- **Alcohol use**: Heavy alcohol consumption is linked to a higher risk of lung cancer, particularly when combined with tobacco smoking.
- **Dust Allergy**: Allergies to dust may exacerbate respiratory symptoms but are not directly linked to lung cancer development.
- **Occupational Hazards**: Exposure to occupational carcinogens such as asbestos, silica, arsenic, and certain chemicals increases the risk of lung cancer.
- **Genetic Risk**: Family history of lung cancer or genetic predispositions can elevate the likelihood of developing the disease.
- **Chronic Lung Disease**: Pre-existing chronic lung conditions like chronic obstructive pulmonary disease (COPD) or emphysema heighten the risk of lung cancer.
- **Balanced Diet**: A diet rich in fruits, vegetables, and antioxidants may lower the risk of lung cancer, while a poor diet may increase susceptibility.
- **Obesity**: Obesity is associated with an increased risk of lung cancer, although the exact mechanism is not fully understood.
- **Smoking**: Tobacco smoking is the leading cause of lung cancer, responsible for the majority of cases worldwide.
- **Passive Smoker**: Exposure to secondhand smoke increases the risk of lung cancer, particularly in nonsmokers.
- **Chest Pain**: Chest pain can be a symptom of lung cancer, especially if it persists or worsens over time.
- **Coughing of Blood**: Hemoptysis (coughing up blood) is a common symptom of advanced lung cancer.
- **Fatigue**: Fatigue is a common symptom of lung cancer, often caused by the disease itself or the side effects of treatment.
- **Weight Loss**: Unexplained weight loss can be a sign of advanced lung cancer or cancer cachexia.
- **Shortness of Breath**: Dyspnea (shortness of breath) is a common symptom of lung cancer, particularly as the disease progresses.
- **Wheezing**: Wheezing can occur in lung cancer patients, especially if tumors obstruct the airways.
- **Swallowing Difficulty**: Difficulty swallowing (dysphagia) may occur if lung cancer spreads to nearby structures or lymph nodes.
- **Clubbing of Finger Nails**: Clubbing of the fingers, where the nails curve excessively, can be a sign of advanced lung cancer.
- Frequent Cold: Frequent respiratory infections or colds may be a symptom of an underlying lung condition, including lung cancer.

- Dry Cough: A persistent dry cough can be a symptom of lung cancer, especially if it lasts for several weeks.
- Snoring: While snoring itself is not directly linked to lung cancer, it may be a sign of obstructive sleep apnea, which has been associated with an increased risk of cancer development.

The target variable is lung cancer level, categorized into three stages: low, medium, and high.

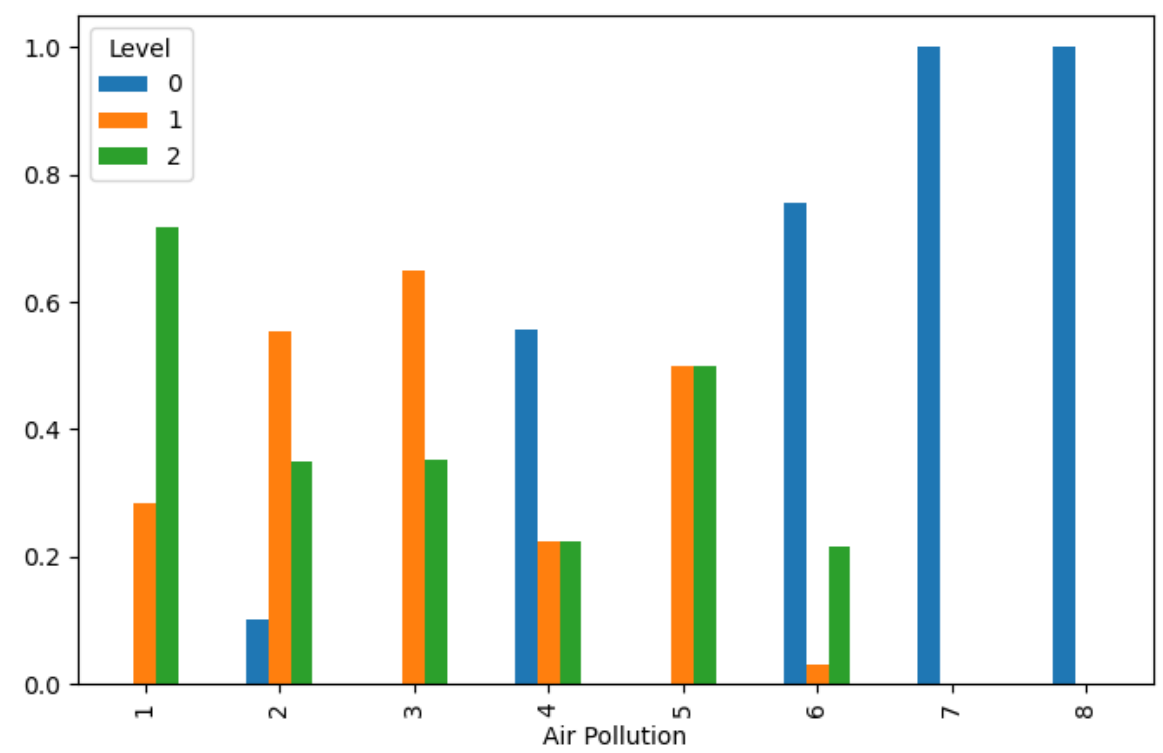

**Figure 1.** Lung Cancer Data, Air Pollution Distribution in hue of Lung Cancer Level.

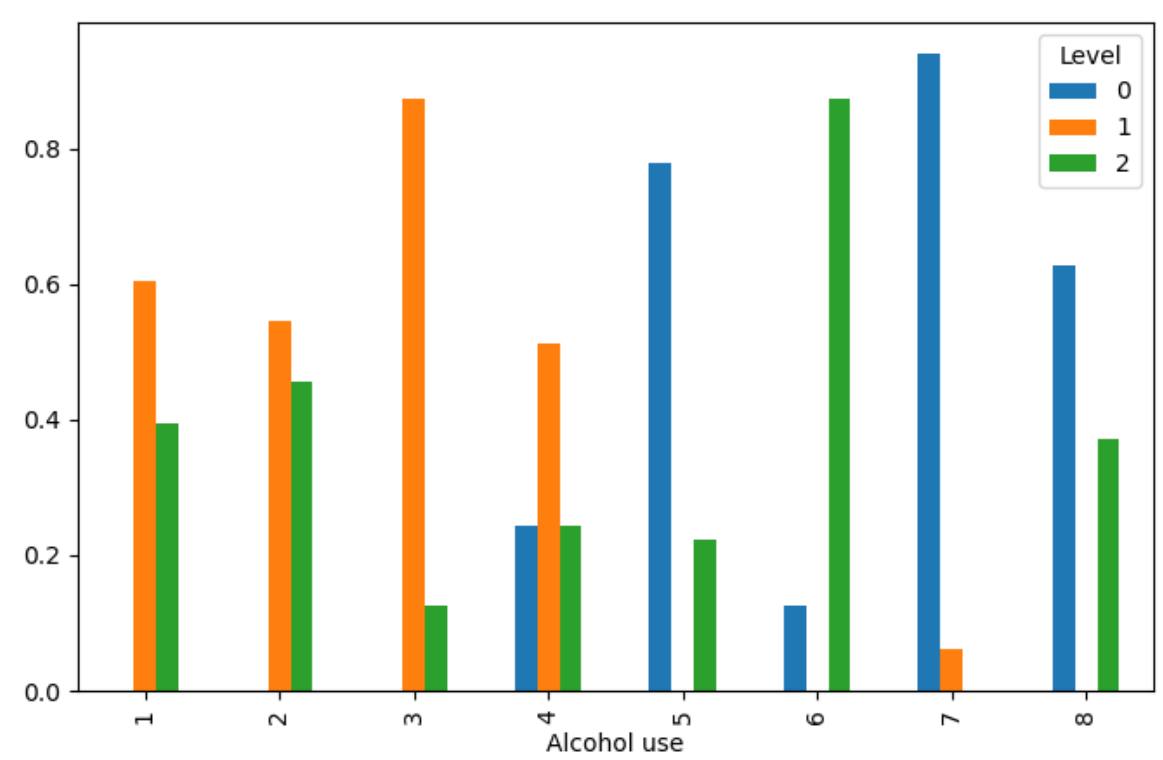

**Figure 2.** Lung Cancer Data, Alcohol Use Distribution in hue of Lung Cancer Level.

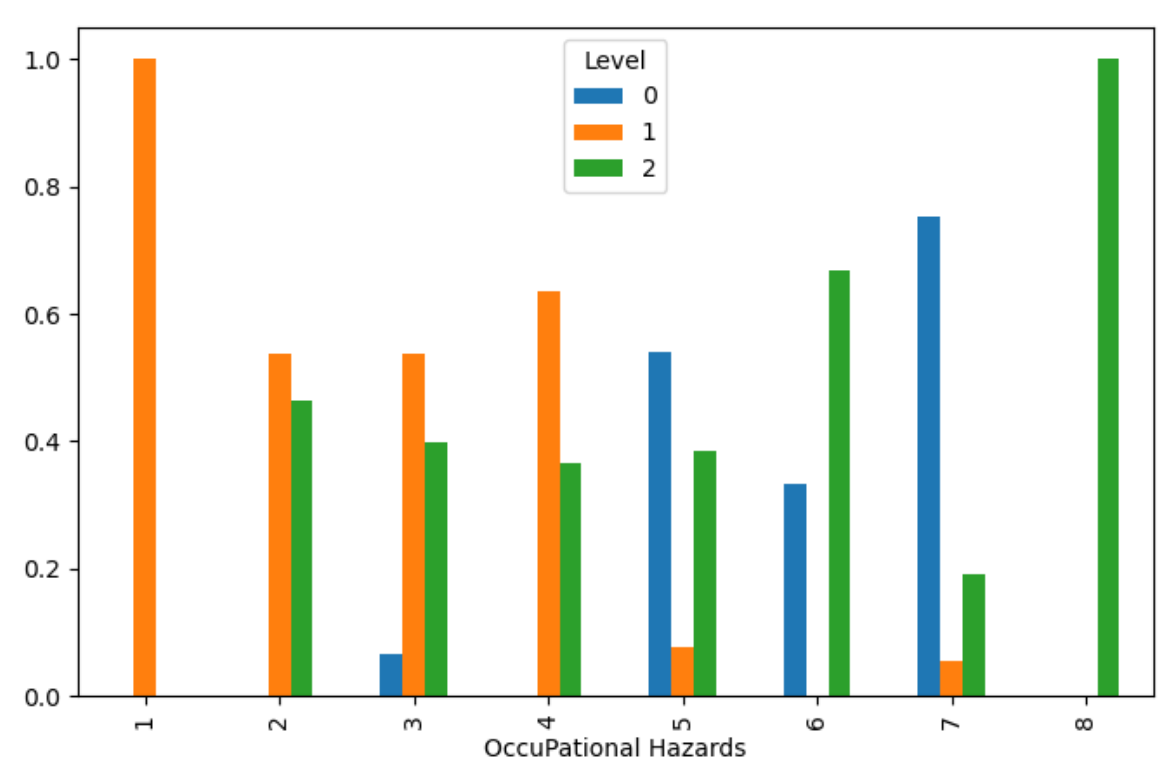

**Figure 3.** Lung Cancer Data, Occupational Hazards Distribution in hue of Lung Cancer Level.

In our methodology section, we delved into the multifaceted relationship between various features and the likelihood of developing lung cancer. Through a comprehensive analysis illustrated in Figures 1-12, we unveiled notable patterns and correlations that shed light on the intricate interplay of factors influencing lung cancer levels. One key observation is the inverse relationship between air pollution levels and lung cancer incidence, as depicted in ***Figure 1***. This finding underscores the critical role of environmental factors in shaping disease prevalence and highlights the potential impact of pollution mitigation efforts on public health outcomes. Furthermore, our analysis revealed a direct correlation between increased occupational hazards and higher lung cancer levels, as evidenced in ***Figure 3***. This highlights the importance of workplace safety measures in reducing occupational health risks and preventing lung cancer among exposed populations. Additionally, genetic predisposition emerged as a significant determinant of lung cancer susceptibility, with higher genetic risk distributions associated with lower cancer levels, as depicted in ***Figure 4***. This underscores the complex interplay between genetic and environmental factors in disease pathogenesis and underscores the importance of personalized risk assessment in clinical practice.

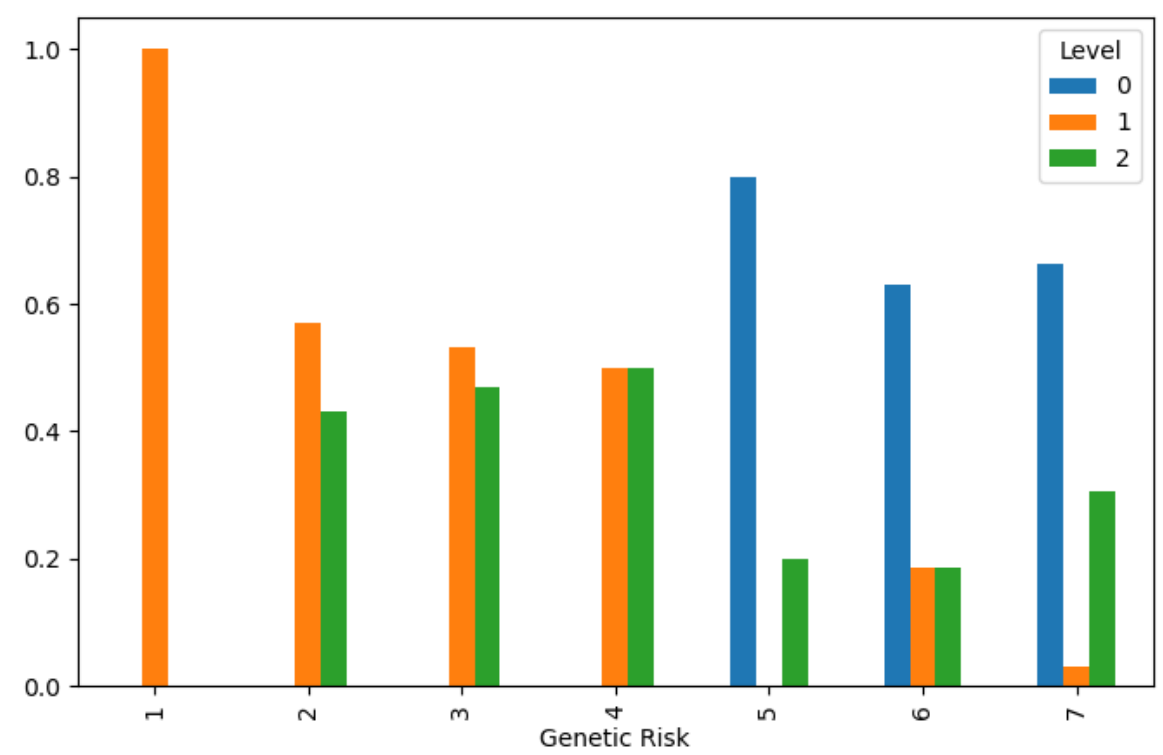

**Figure 4.** Lung Cancer Data, Genetic Risk Distribution in hue of Lung Cancer Level.

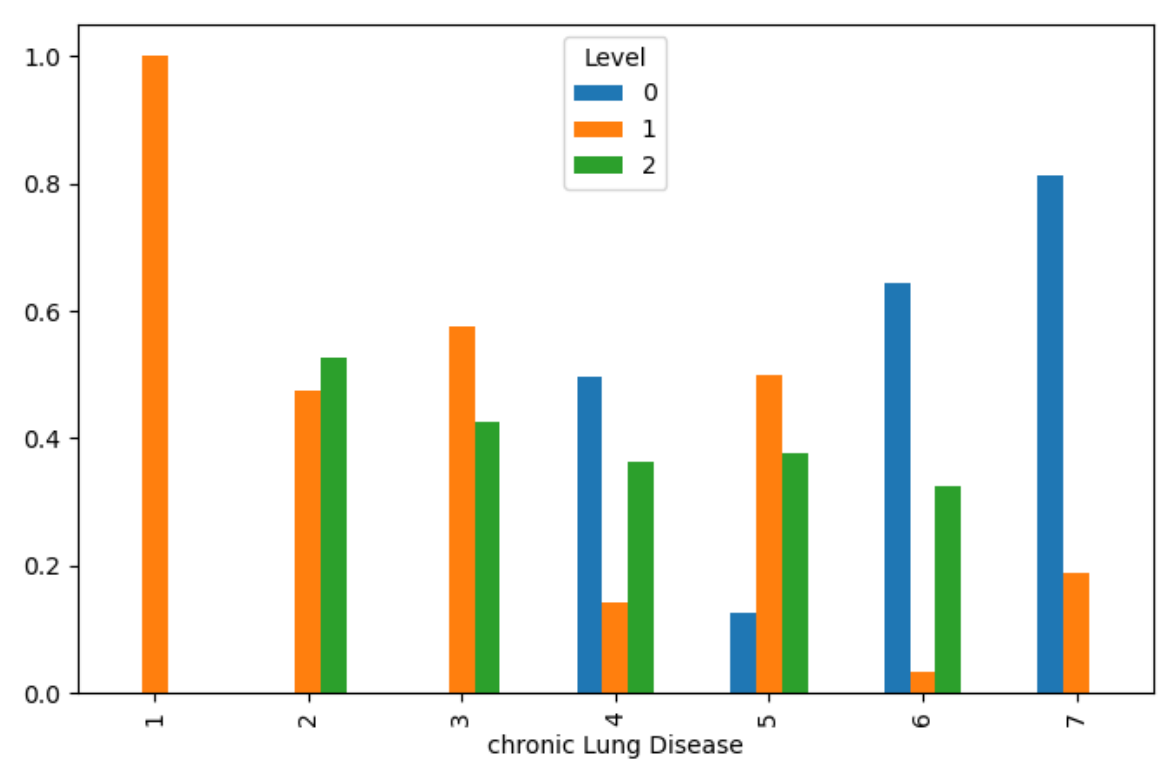

**Figure 5.** Lung Cancer Data, Chronic Lung Disease Distribution in hue of Lung Cancer Level.

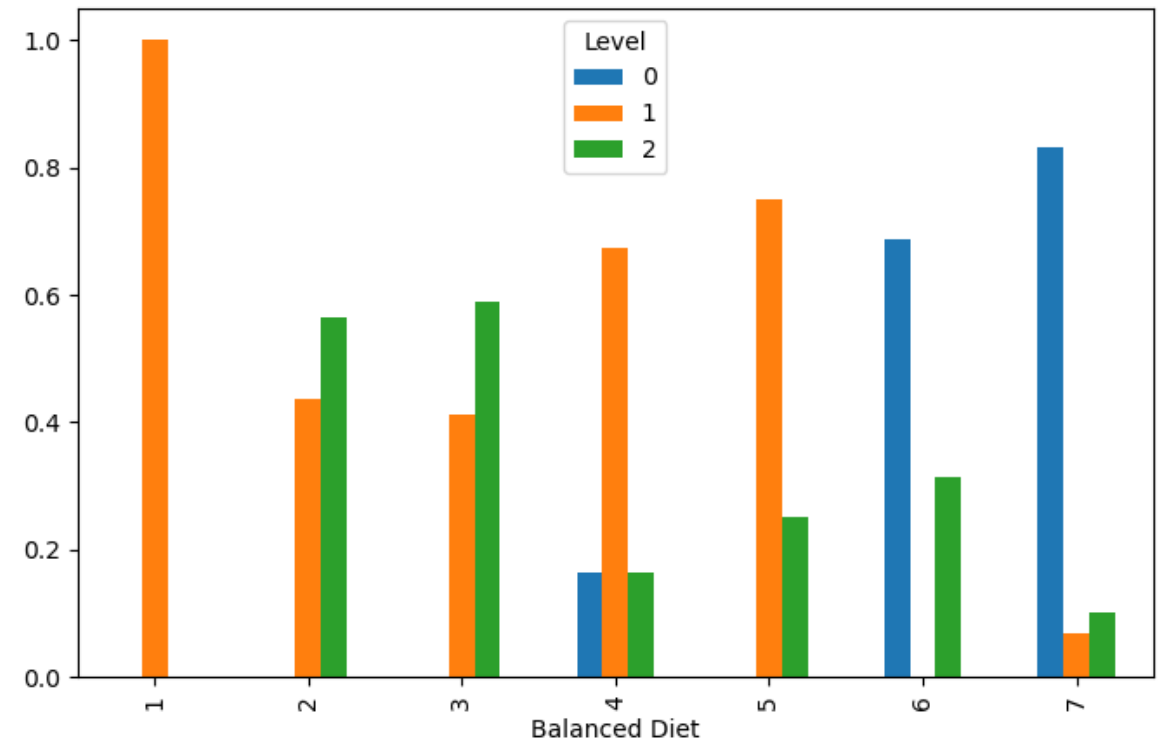

**Figure 6.** Lung Cancer Data, Balanced Diet Distribution in hue of Lung Cancer Level.

Moreover, our analysis uncovered intriguing associations between lifestyle factors and lung cancer incidence. Chronic lung disease and a balanced diet distribution were found to correspond with lower cancer levels, as illustrated in ***Figure 5*** and ***Figure 6***, respectively. These findings underscore the potential of lifestyle modifications and preventive interventions in reducing lung cancer risk and improving public health outcomes. However, the effects of obesity and smoking distribution on lung cancer incidence were more nuanced, with varied impacts leading to both low and high cancer levels, as depicted in ***Figure** 7* and ***Figure 8***. These findings underscore the complex interplay of behavioral and metabolic factors in lung cancer etiology and highlight the importance of targeted interventions in high-risk populations.

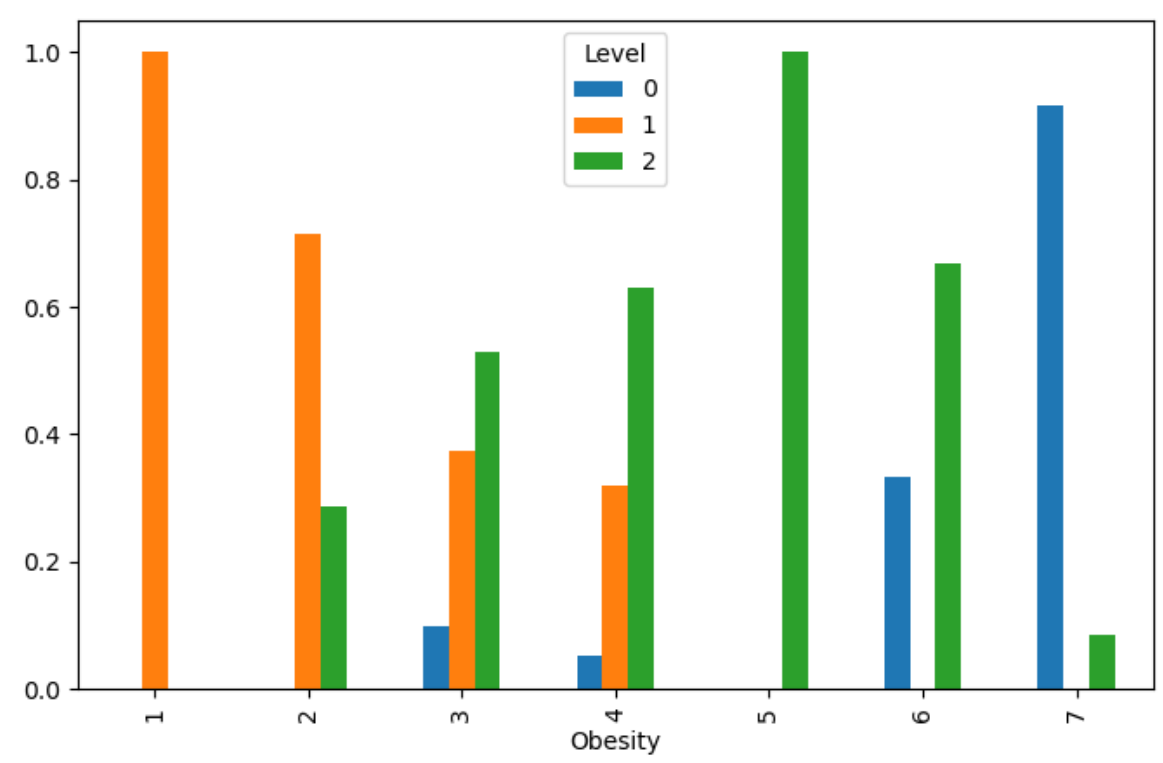

**Figure 7.** Lung Cancer Data, Obesity Distribution in hue of Lung Cancer Level.

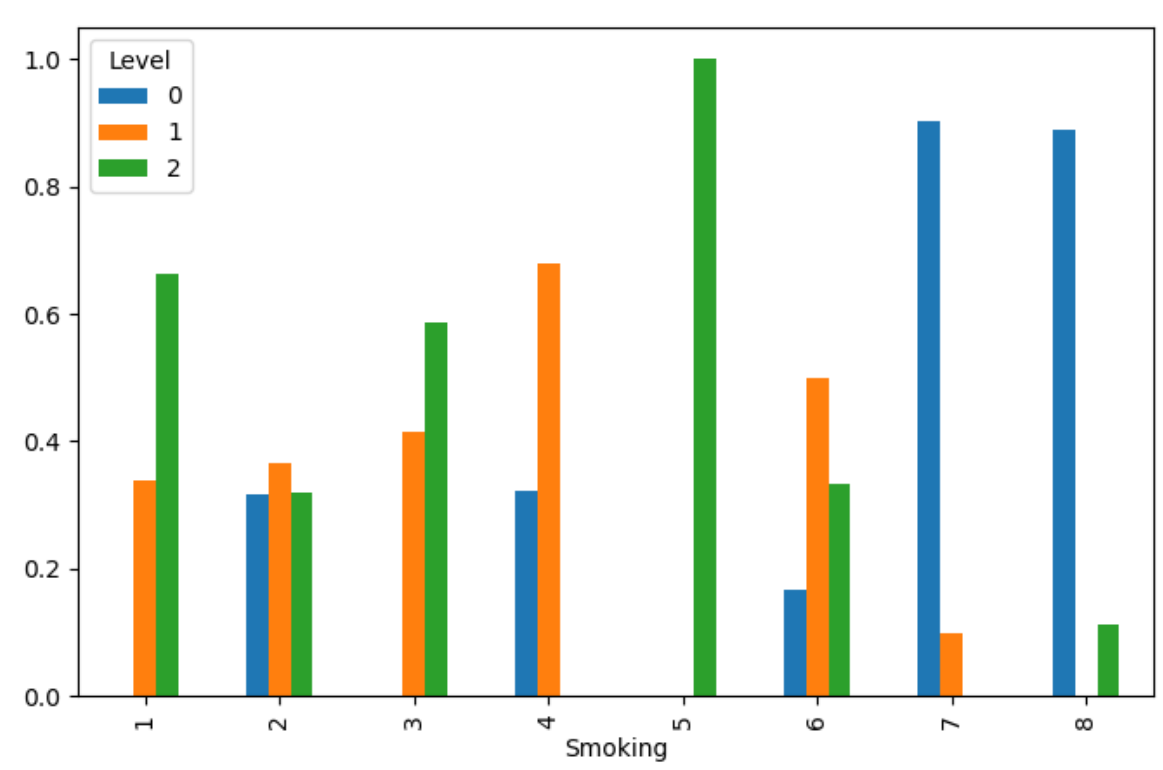

**Figure 8.** Lung Cancer Data, Smoking Distribution in hue of Lung Cancer Level.

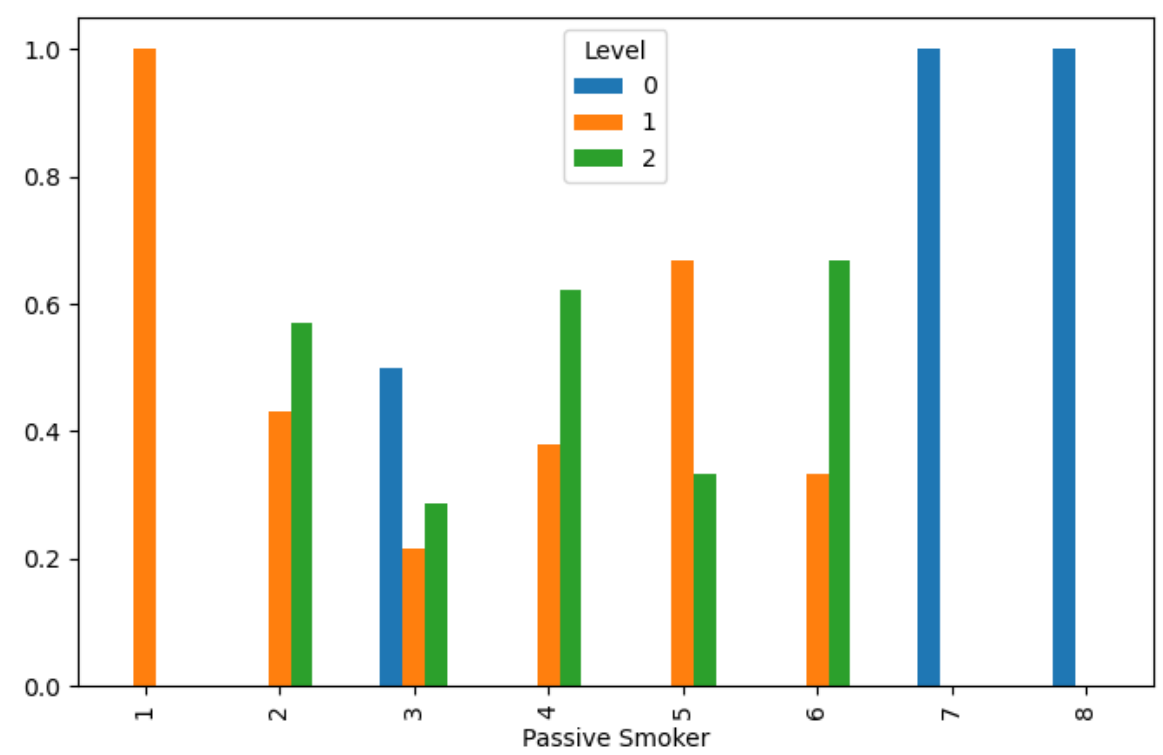

**Figure 9.** Lung Cancer Data, Passive Smoker Distribution in hue of Lung Cancer Level.

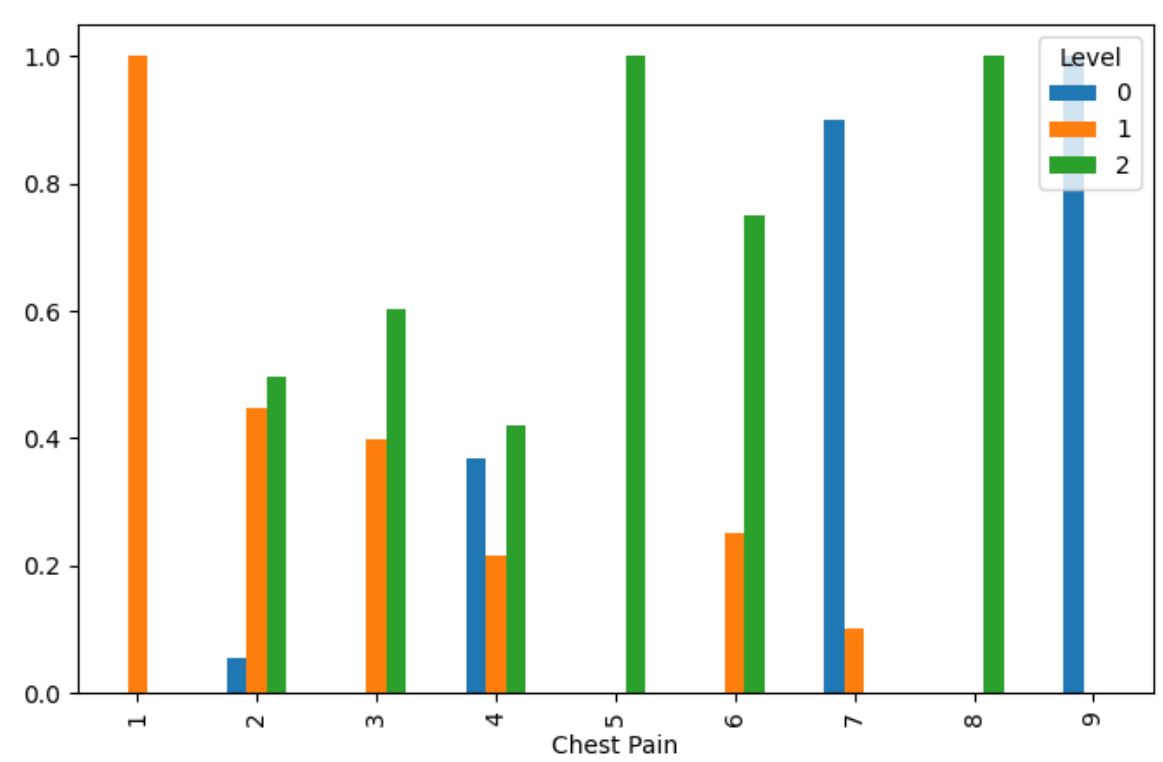

**Figure 10.** Lung Cancer Data, Chest Pain Distribution in hue of Lung Cancer Level.

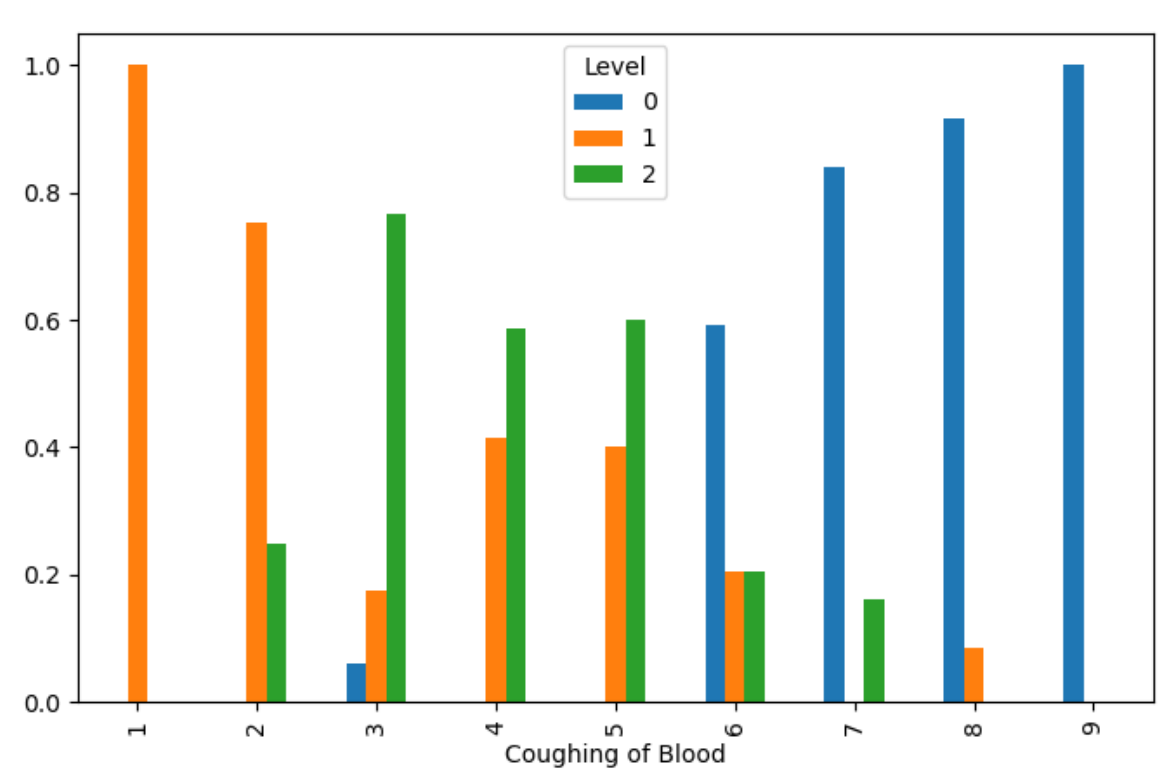

**Figure 11.** Lung Cancer Data, Coughing of Blood Distribution in hue of Lung Cancer Level.

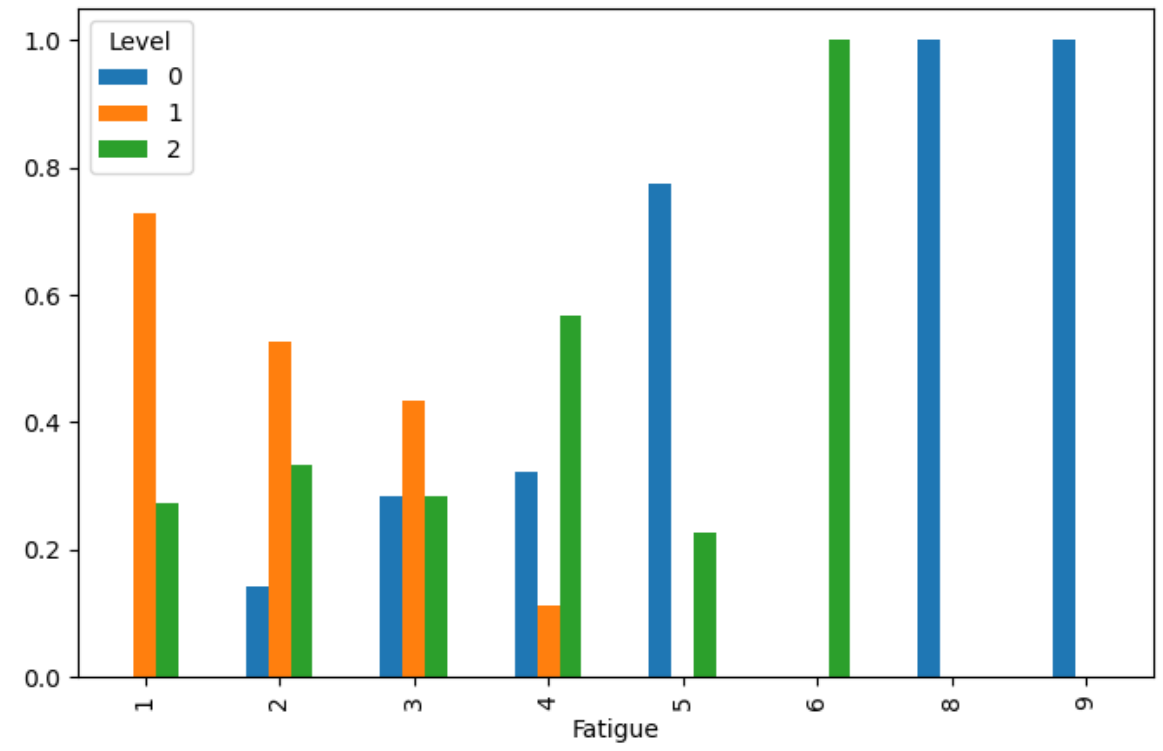

**Figure 12.** Lung Cancer Data, Fatigue Distribution in hue of Lung Cancer Level.

correlation analysis emerged as a pivotal tool in discerning meaningful relationships between features within our dataset. By employing correlation plots, we systematically identified associations with potential significance for our analysis. Features exhibiting a correlation coefficient exceeding 0.5 were deemed pertinent for further exploration. These high correlations prompted the creation of new data frame columns through feature combination, a strategic approach aimed at enhancing the predictive efficacy of our models. This process of feature engineering served to enrich the dataset, enabling our models to capture nuanced relationships and derive more accurate predictions. Conversely, features displaying a correlation coefficient lower than -0.4 were considered to lack meaningful correlation within the context of our analysis. While these findings did not diminish the importance of these features outright, they underscored the need for cautious interpretation and prioritization in subsequent modeling efforts.

This methodological approach, rooted in rigorous statistical analysis, forms a cornerstone of our study's robustness and integrity. By systematically identifying and leveraging meaningful correlations, we strive to enhance the effectiveness and reliability of our predictive models, ultimately contributing to advancements in the field of data-driven decision-making. Overall, our methodology section provides valuable insights into the multifactorial nature of lung cancer etiology, highlighting the diverse array of factors influencing disease risk and progression. By elucidating these complex relationships, our analysis lays the groundwork for future research endeavors aimed at improving risk prediction, early detection, and prevention strategies for lung cancer.

## Splitting Data

In the Data Splitting section, a rigorous approach was adopted to ensure the integrity and generalizability of our machine learning models. The datasets underwent meticulous partitioning into training and test sets, a critical step aimed at mitigating overfitting and maintaining data balance. To address potential disparities in feature scales and class distribution, two key preprocessing techniques were employed. Firstly, the MinMaxScaler was utilized for data normalization, ensuring uniformity in feature magnitudes and facilitating convergence during model training. Additionally, the Synthetic Minority Over-sampling Technique (SMOTE) from imblearn.over_sampling was applied to address class imbalance, thereby promoting equitable representation of minority classes in the training data. Following data preprocessing, the train_test_split method was invoked, resulting in a stratified partitioning of 876 samples for training and 219 samples for testing, as illustrated in Figure 13. This stratification ensures that the distribution of classes remains consistent between the training and test sets, thereby enabling robust model evaluation and performance assessment.

Furthermore, to ascertain the stability and reliability of our models, k-fold cross-validation was employed with k=5. This technique partitions the dataset into k equal-sized folds, with each fold serving alternately as the validation set while the remaining folds are used for training. By iteratively training and evaluating the model across different subsets of the data, k-fold cross-validation aids in mitigating overfitting and provides a more accurate estimation of model performance.

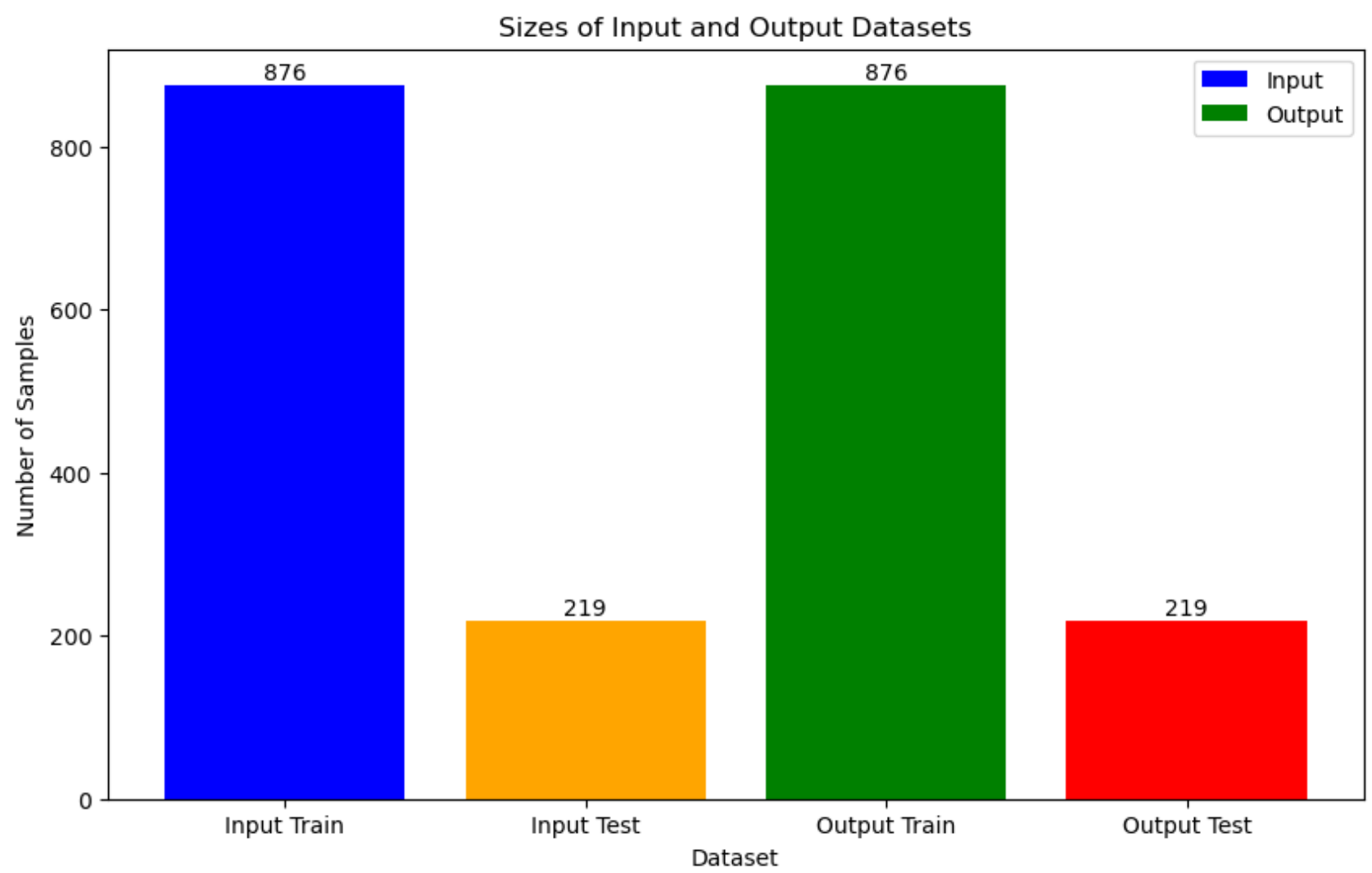

**Figure 13.** Splitting Lung Cancer Data into Training and Test Datasets.

## ML Models

Six machine learning models were employed for lung cancer prediction:

- Deep Neural Network (DNN)
- Voting
- Bagging
- Support Vector Machine (SVC) with radial basis function kernel (SVC_rbf)
- SVC with linear kernel (SVC_linear)
- SVC with polynomial kernel (SVC_polynomial)
- SVC with sigmoid kernel (SVC_sigmoid)

Each of these models was carefully selected to ensure comprehensive coverage and robust prediction capabilities.

### Deep Neural Networks (DNN)

Deep Neural Networks (DNNs) represent a class of artificial neural networks distinguished by their numerous hidden layers positioned between the input and output layers. These networks excel at capturing intricate patterns and correlations within data, particularly in scenarios involving high-dimensional input data such as images, audio, and text [13]. Leveraging deep architectures, DNNs autonomously learn hierarchical representations of features, enabling them to effectively discern between various classes within the dataset [4, 5, 15, 16].

### Voting Classifier

The Voting Classifier operates by amalgamating predictions from multiple individual classifiers to produce a final prediction. It consolidates the decisions made by each classifier and selects the class with the most votes as the predicted outcome. This ensemble approach typically yields enhanced performance compared to individual classifiers by leveraging the diverse viewpoints provided by multiple models.

### Bagging (Bootstrap Aggregating)

Bagging, short for Bootstrap Aggregating, is an ensemble learning technique that entails training multiple instances of the same base classifier on different subsets of the training data. The final prediction is usually derived by averaging the predictions generated by all individual classifiers. Bagging serves to diminish variance and mitigate overfitting by introducing randomness into the training process and fostering model diversity [17]. Support Vector Machines (SVMs) are potent supervised learning models utilized for classification tasks. The RBF kernel is a favored choice for SVM classification due to its capability to capture intricate nonlinear relationships among features. SVMs strive to ascertain the hyperplane that most effectively separates the classes in the feature space, endeavoring to maximize the margin between distinct classes while minimizing classification errors.

### Support Vector Machine (SVM) with Linear Kernel (SVC_linear)

The linear kernel represents another variant of SVM that assumes the input data to be linearly separable. It functions by identifying the optimal linear boundary (hyperplane) that segregates the classes in the feature space. Despite its simplicity, linear SVMs can exhibit commendable performance across many classification tasks, particularly when the data displays linear separability or when the feature space is high-dimensional.

### Support Vector Machine (SVM) with Polynomial Kernel (SVC_polynomial)

The polynomial kernel is employed in SVMs to handle nonlinear relationships among features by projecting the input data into a higher-dimensional space. This facilitates SVMs in capturing more intricate decision boundaries compared to linear SVMs. However, selecting an appropriate degree for the polynomial kernel is imperative to forestall overfitting.

### Support Vector Machine (SVM) with Sigmoid Kernel (SVC_sigmoid)

The sigmoid kernel serves as an alternative option for SVM classification, particularly suitable for binary classification tasks. It operates akin to logistic regression and can delineate nonlinear decision boundaries. Nonetheless, SVMs with sigmoid kernels may exhibit sensitivity to hyperparameter settings and susceptibility to overfitting, necessitating meticulous tuning for optimal performance.

## Results and Discussion

Overfitting is a common challenge in machine learning, occurring when a model learns to perform well on the training data but fails to generalize to unseen data. In the context of lung cancer level classification, overfitting can lead to inaccurate predictions and unreliable model performance, ultimately hindering the effectiveness of the classification system.

There are several reasons why overfitting occurs and why it needs to be addressed in the context of lung cancer level classification:

- Complexity of Data: Lung cancer classification involves intricate relationships between various features such as patient characteristics, environmental factors, and medical history. Overfitting can occur when the model captures noise or irrelevant patterns in the training data, leading to poor generalization to new data.

- Limited Sample Size: Medical datasets, including those used for lung cancer classification, are often limited in size due to data collection constraints. With a small dataset, there is a higher risk of overfitting as the model may memorize the training samples rather than learning meaningful patterns.
- Imbalanced Data: Imbalanced classes, such as uneven distribution of lung cancer levels, can exacerbate the overfitting problem. Models may become biased towards the majority class, leading to poor performance in predicting minority classes.

To mitigate the risk of overfitting in lung cancer level classification, it is essential to apply techniques such as adjusting hyperparameters, including learning rate and minimum child weight. Monitoring these parameters allows us to control the complexity of the model and prevent it from memorizing noise in the training data. By fine-tuning hyperparameters, we aim to strike a balance between model complexity and generalization performance, ensuring that the model can effectively classify lung cancer levels in unseen data.

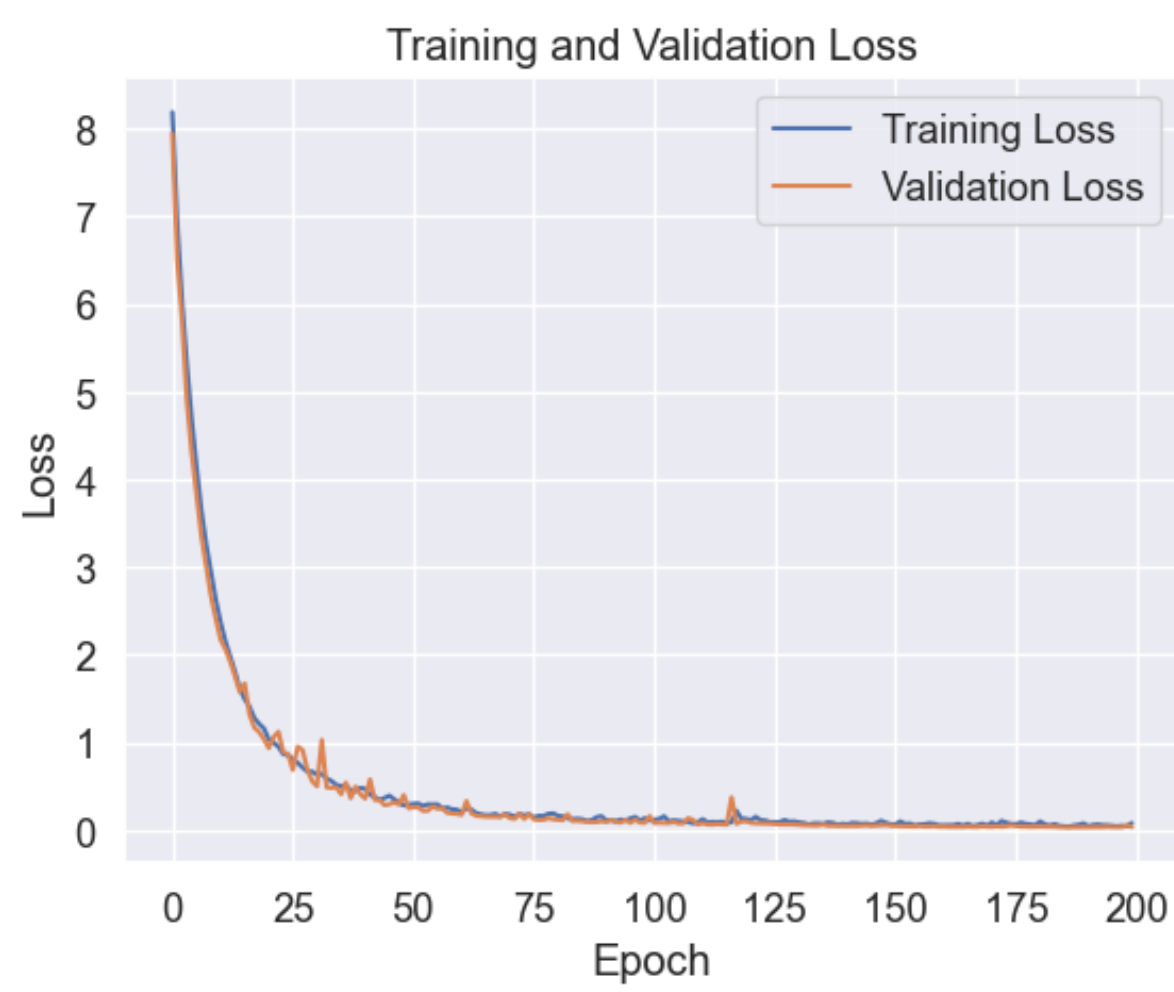

**Figure 14.** Training vs. Validation in DNN.

In the Results and Discussion section, we unveil the insights gleaned from our comprehensive analysis of machine learning models for lung cancer level classification. Through meticulous parameter monitoring and optimization, we endeavored to enhance the accuracy and reliability of our classification model while mitigating the risk of overfitting. Of the six ML models scrutinized in our study, the Deep Neural Network (DNN) emerged as a standout performer, showcasing robust performance across various evaluation metrics. Notably, our analysis of the learning plot in ***Figure 14*** revealed a remarkable alignment between the training and validation errors, indicative of minimal overfitting. This finding underscores the efficacy of our parameter tuning efforts in promoting model generalizability and performance consistency across different datasets.

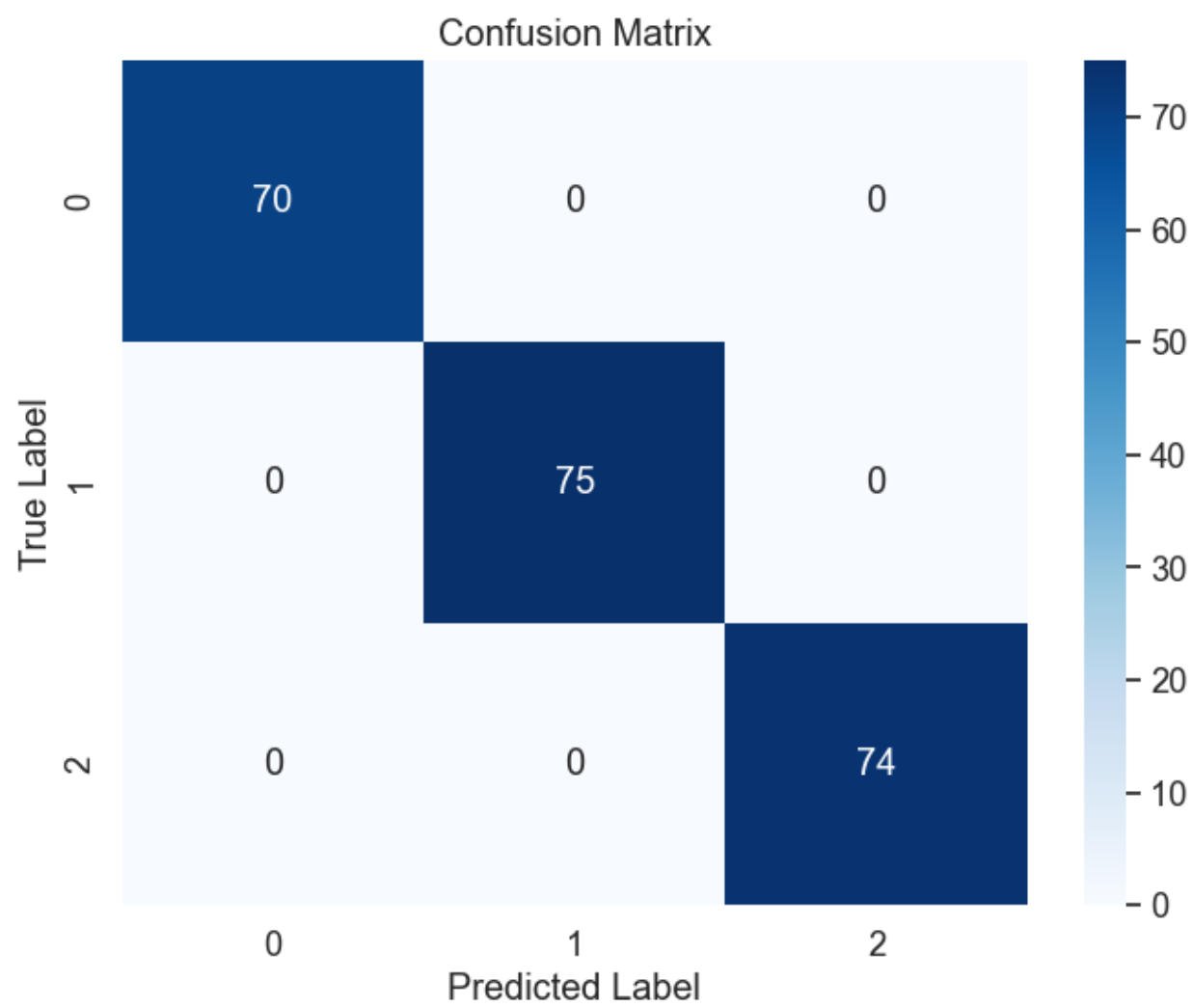

**Figure 15.** Confusion Matrix of DNN.

Upon testing the DNN model with unseen datasets, our findings surpassed expectations, with the model achieving near-perfect accuracy, precision, recall, and F-1 score metrics, as evidenced by the confusion matrix in ***Figure 15***. This flawless performance underscores the DNN's prowess in accurately predicting lung cancer levels, offering promising implications for clinical practice and oncological diagnostics. Furthermore, our study sheds light on the broader implications of our findings for lung cancer diagnosis and treatment planning. By harnessing the predictive power of machine learning models like the DNN, clinicians can leverage advanced analytics to augment their decision-making processes and improve patient outcomes. Additionally, the robust performance of the DNN underscores the potential for integrating such models into clinical workflows, paving the way for more personalized and effective approaches to lung cancer management. Furthermore, our investigation extended to the utilization of two widely recognized ensemble methods, namely voting and bagging, for the classification of lung cancer levels, as illustrated in Figures 16-21. These ensemble techniques offer a synergistic approach to model aggregation, leveraging the collective wisdom of multiple base classifiers to enhance predictive performance. In our analysis, the voting model exhibited commendable stability, as evidenced by the consistent performance across varying learning rates and minimum child weights, as depicted in ***Figure 16***. This robustness underscores the reliability of the ensemble approach in mitigating the effects of individual model variability and promoting consistent predictions across different parameter settings.

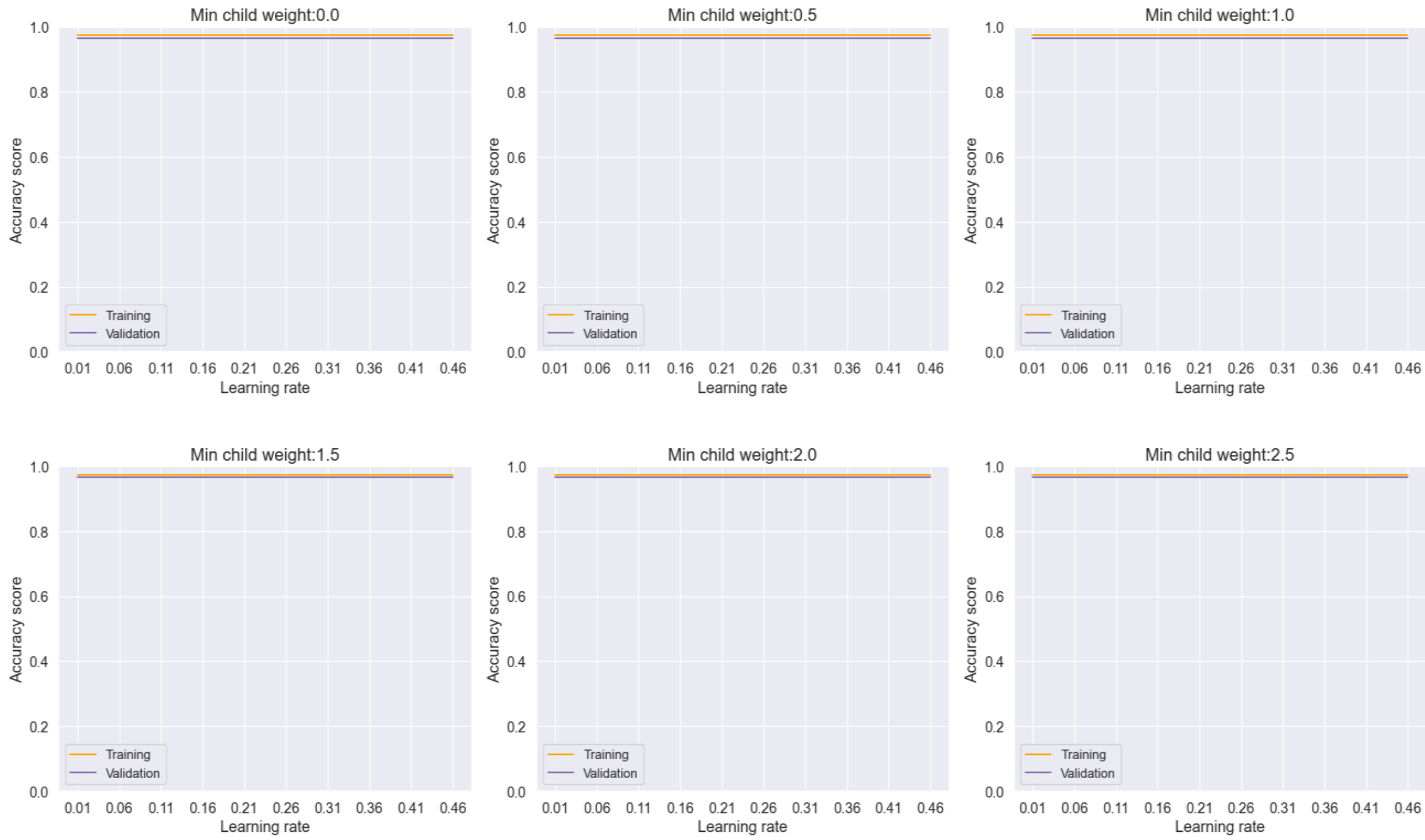

**Figure 16.** Training and Validation plots under consideration of Min child Weight and Learning Rate in Ensemble Methods, Voting.

Subsequent evaluation using unseen data further corroborated the effectiveness of the voting model, with the absence of errors evident in the confusion matrix, as illustrated in ***Figure 17***. This outcome attests to the model's ability to generalize well to new data, a crucial consideration in real-world applications where model performance on unseen samples is paramount. Moreover, the learning curve analysis provided additional insights into the convergence behavior of the voting model, revealing convergence at the highest score, as depicted ***Figure 18***. This convergence underscores the model's capacity to attain optimal performance with sufficient training data, further reinforcing its utility in lung cancer level classification tasks.

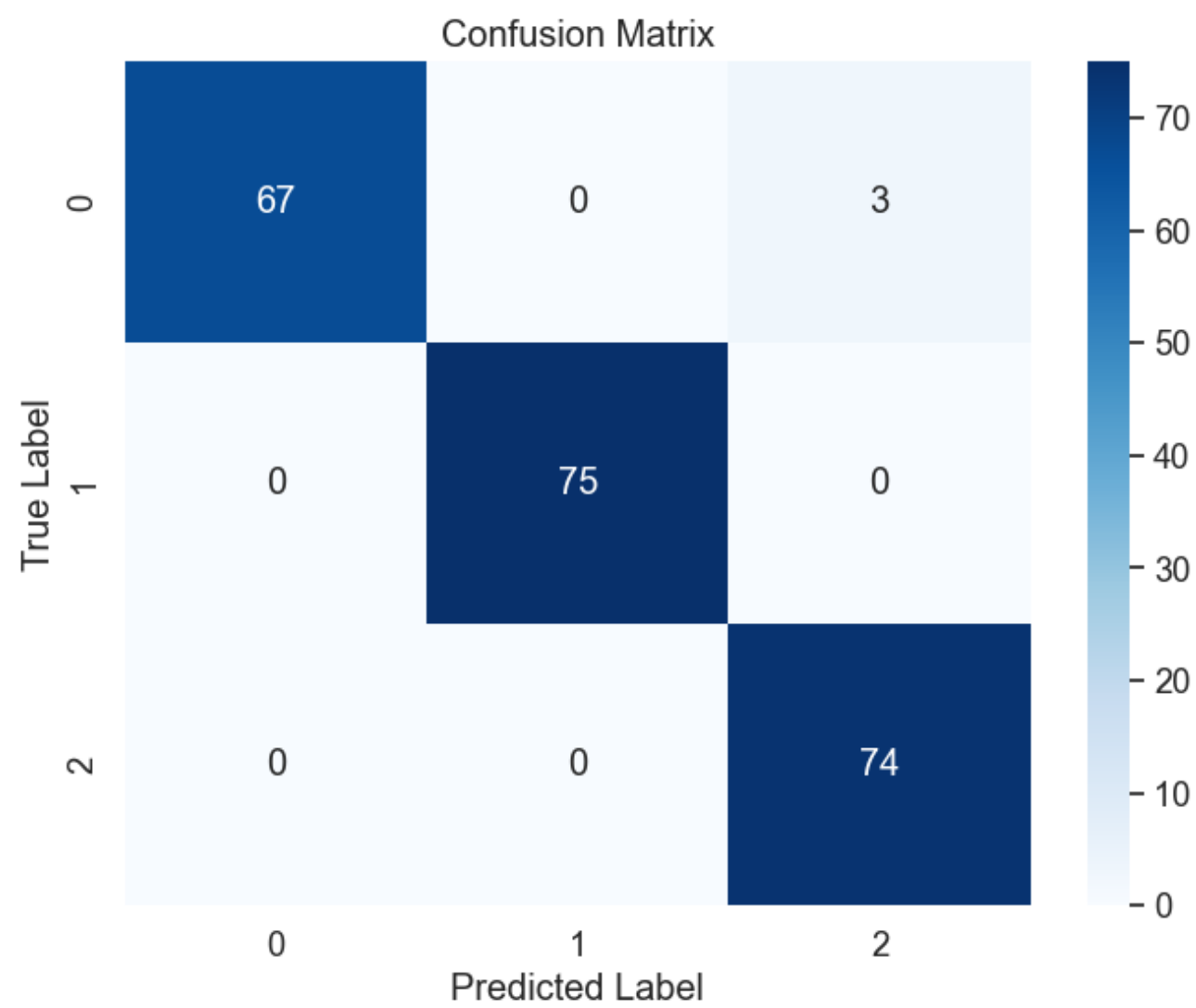

**Figure 17.** Confusion Matrix of Ensemble Methods, Voting.

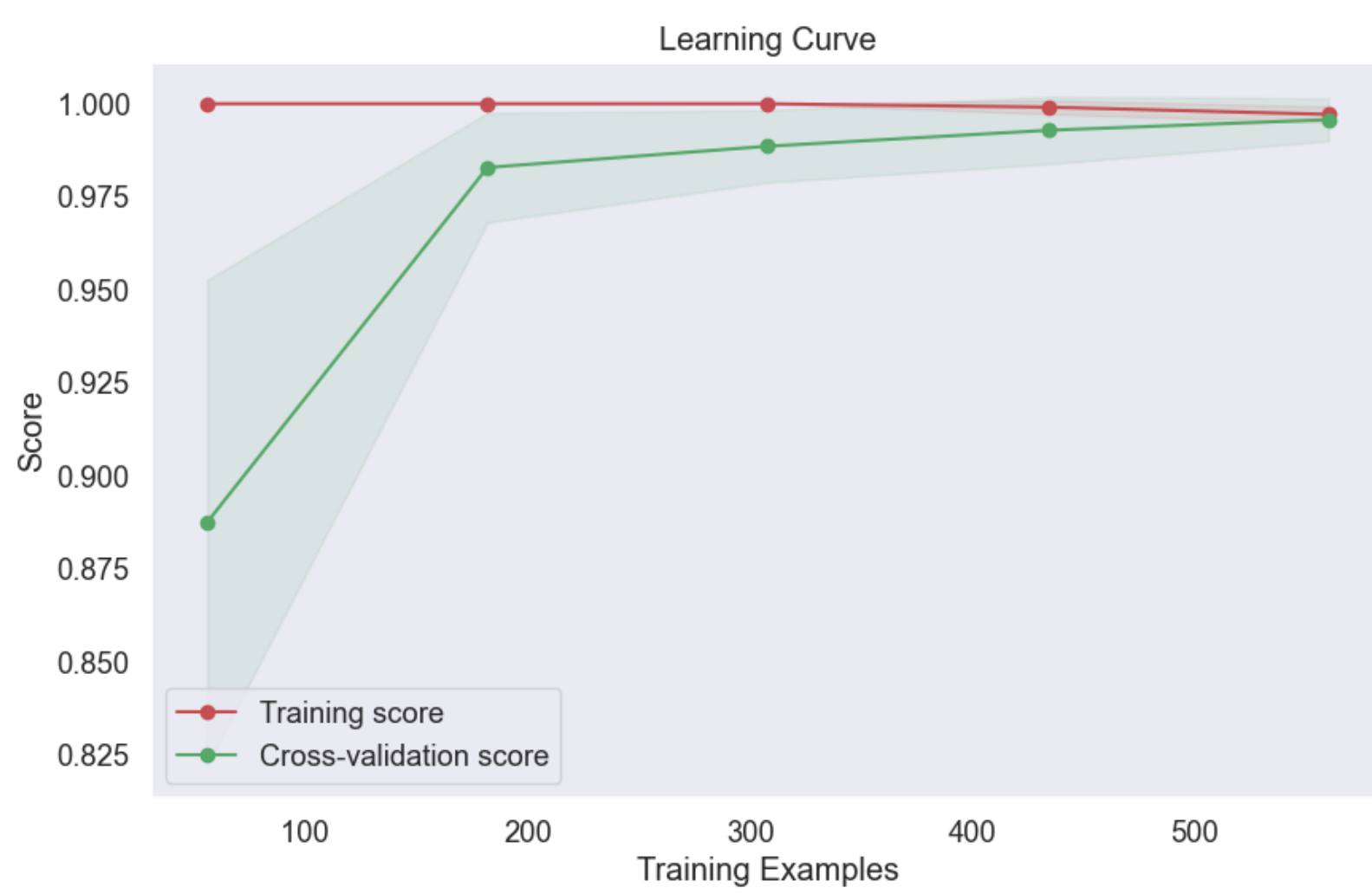

**Figure 18.** Learning Curve of Ensemble Methods, Voting.

Likewise, our investigation revealed the remarkable performance of the bagging model across all levels of lung cancer datasets, as depicted in ***Figure 19***. The exemplary outcomes were further substantiated by the analysis of the confusion matrix, as illustrated in ***Figure 20***, where the absence of errors underscores the model's ability to accurately classify lung cancer levels with precision and reliability.

Furthermore, the learning curve analysis presented in ***Figure 21*** provided valuable insights into the convergence behavior of the bagging model. The consistent upward trajectory of the learning curve indicates steady improvement in model performance with increasing training data, culminating in optimal performance at convergence. This robust convergence underscores the model's capacity to leverage diverse training samples effectively, thereby enhancing its predictive capabilities and generalization to unseen data. The exceptional performance of the bagging model underscores the efficacy of ensemble techniques in lung cancer level classification. By aggregating predictions from multiple base classifiers, the bagging model demonstrates robustness, stability, and superior predictive accuracy, offering promising avenues for enhancing oncological diagnostics and patient care outcomes.

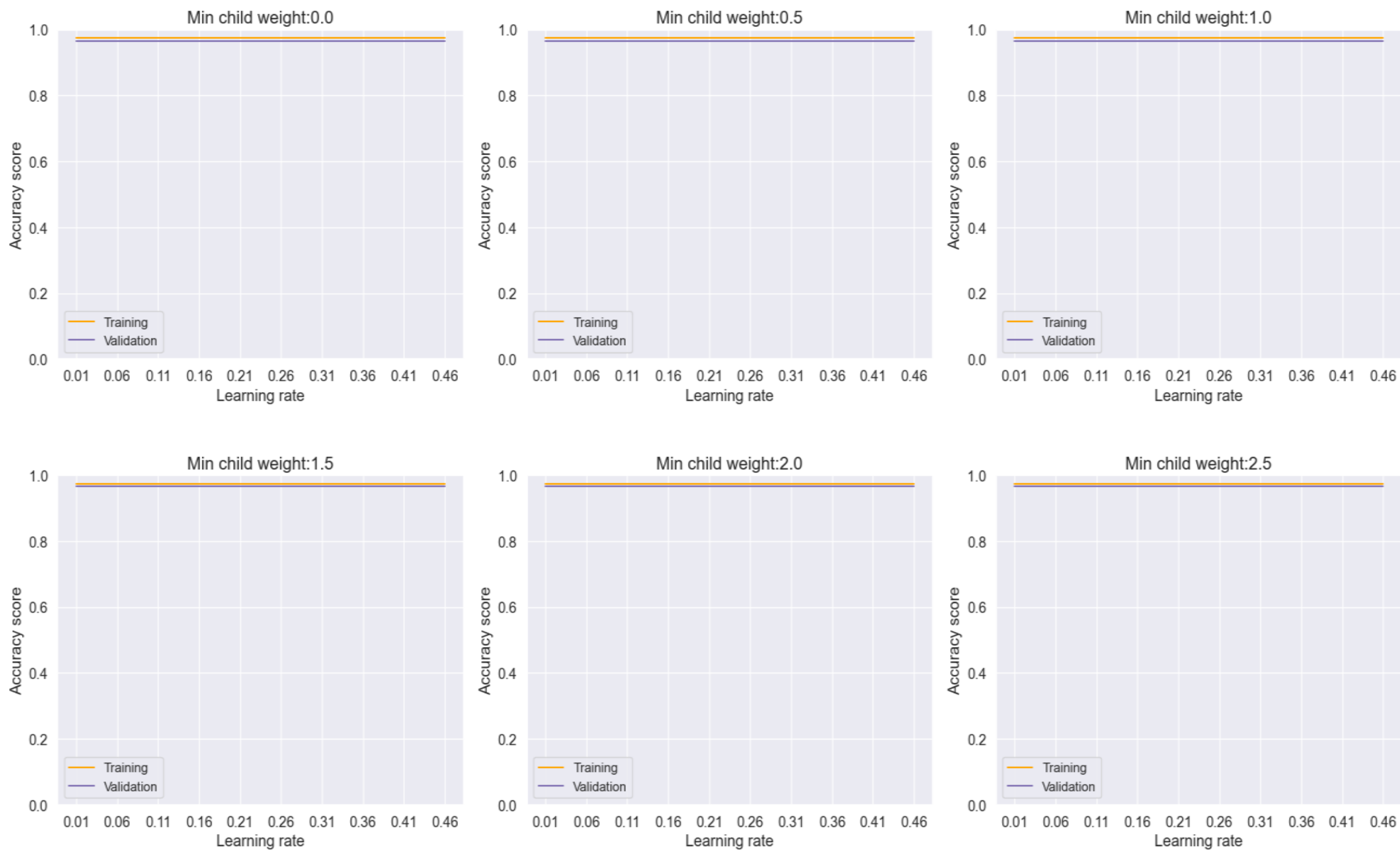

**Figure 19.** Training and Validation plots under consideration of Min child Weight and Learning Rate in Ensemble Methods, Bagging.

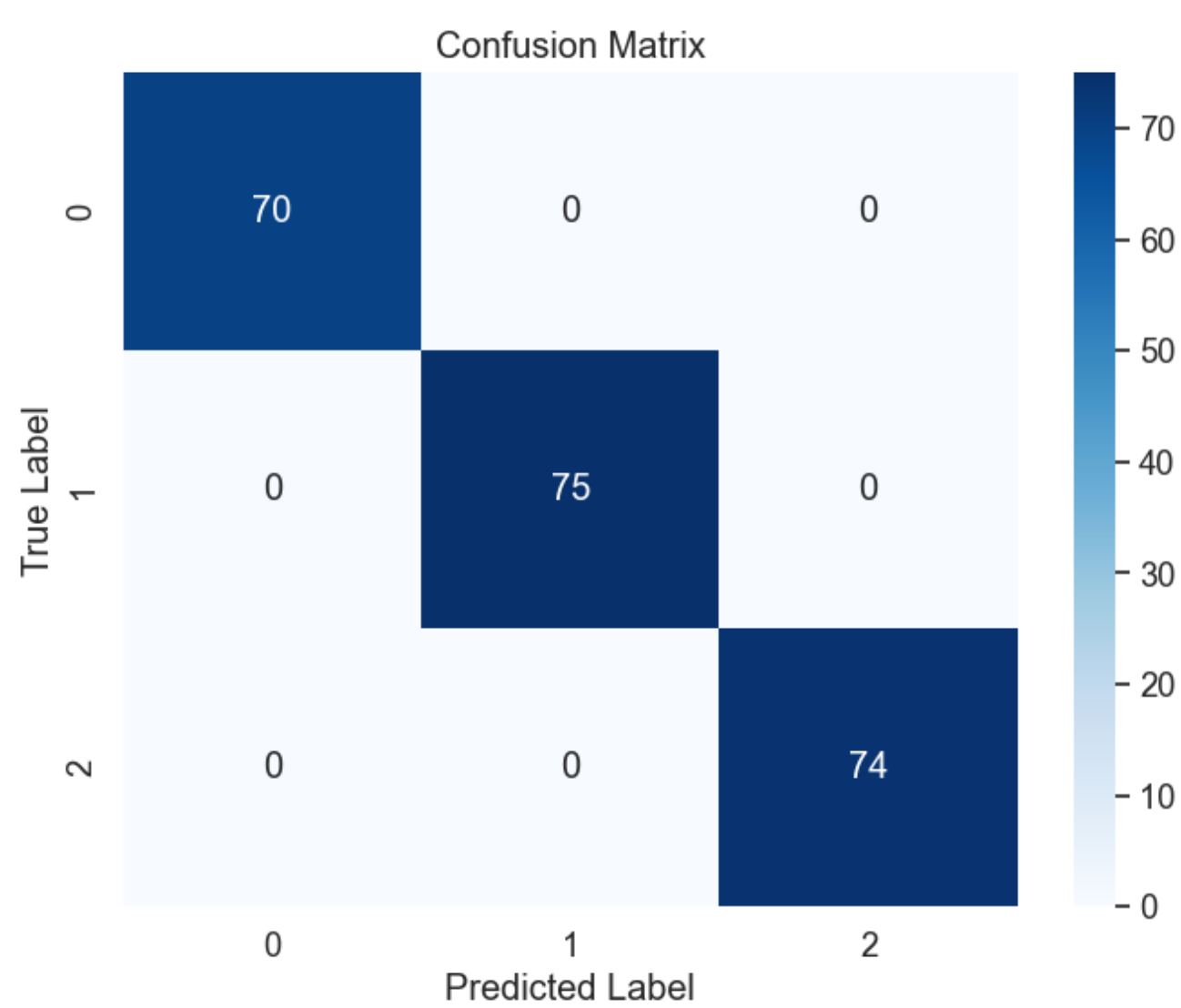

**Figure 20.** Confusion Matrix of Ensemble Methods, Bagging.

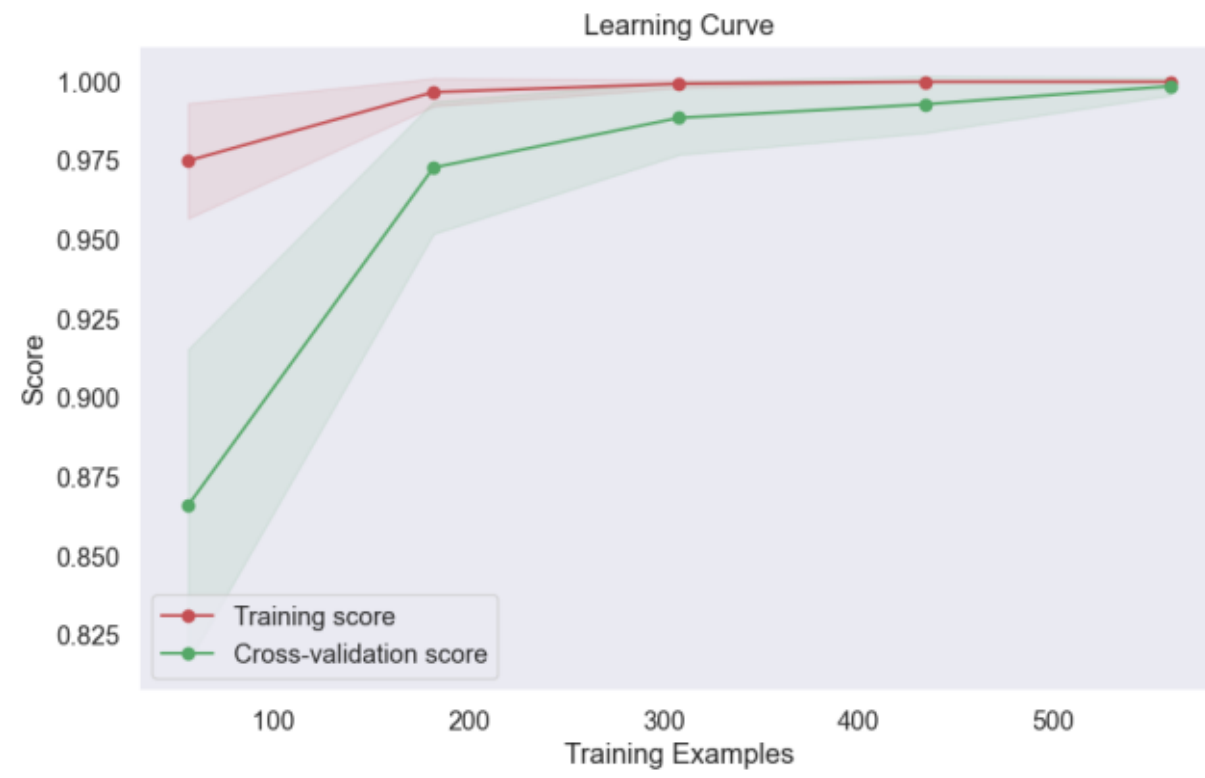

**Figure 21.** Learning Curve of Ensemble Methods, Bagging.

In contrast, our investigation extended to the exploration of Support Vector Machine (SVM) models employing various kernels, including RBF (***Figure 22***, ***Figure 23***)), Linear (***Figure 24***, ***Figure 25***), Polynomial (***Figure 26***, ***Figure 27***), and Sigmoid (***Figure 28***, ***Figure 29***). The objective was to uncover any underlying complexities within the dataset and assess the efficacy of SVM-based approaches in lung cancer level classification. Despite the overall exceptional performance observed with previous models, the SVM model with the Sigmoid kernel encountered significant challenges. This particular model yielded only 37% accuracy, precision, recall, and F-1 score, as highlighted in our analysis. The confusion matrix (***Figure 28***)) revealed a notable prevalence of errors, indicative of the model's limited ability to accurately classify lung cancer levels using the Sigmoid kernel. Moreover, the learning curve analysis provided further insights into the performance limitations of the SVM model with the Sigmoid kernel. ***Figure 29*** illustrates a concerning trend of decreasing accuracy with the addition of more datasets, suggesting a lack of robustness and generalization capability in the face of increased data complexity.

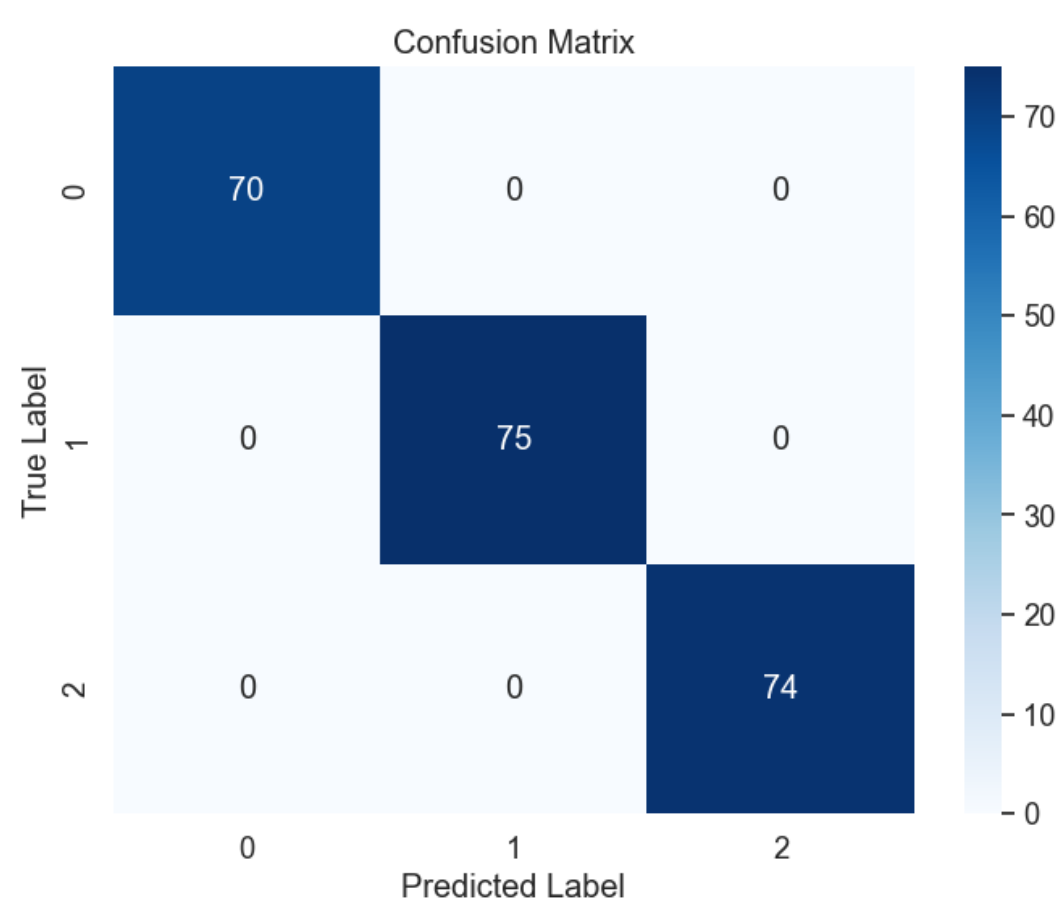

**Figure 22.** Confusion Matrix of SVM with RBF Kernel.

These findings underscore the importance of careful consideration when selecting model architectures and parameters, particularly in complex classification tasks such as lung cancer level classification. While SVM models have demonstrated utility in various machine learning applications, our results highlight the need for caution when employing certain kernel functions, such as the Sigmoid kernel, in scenarios characterized by intricate data relationships.

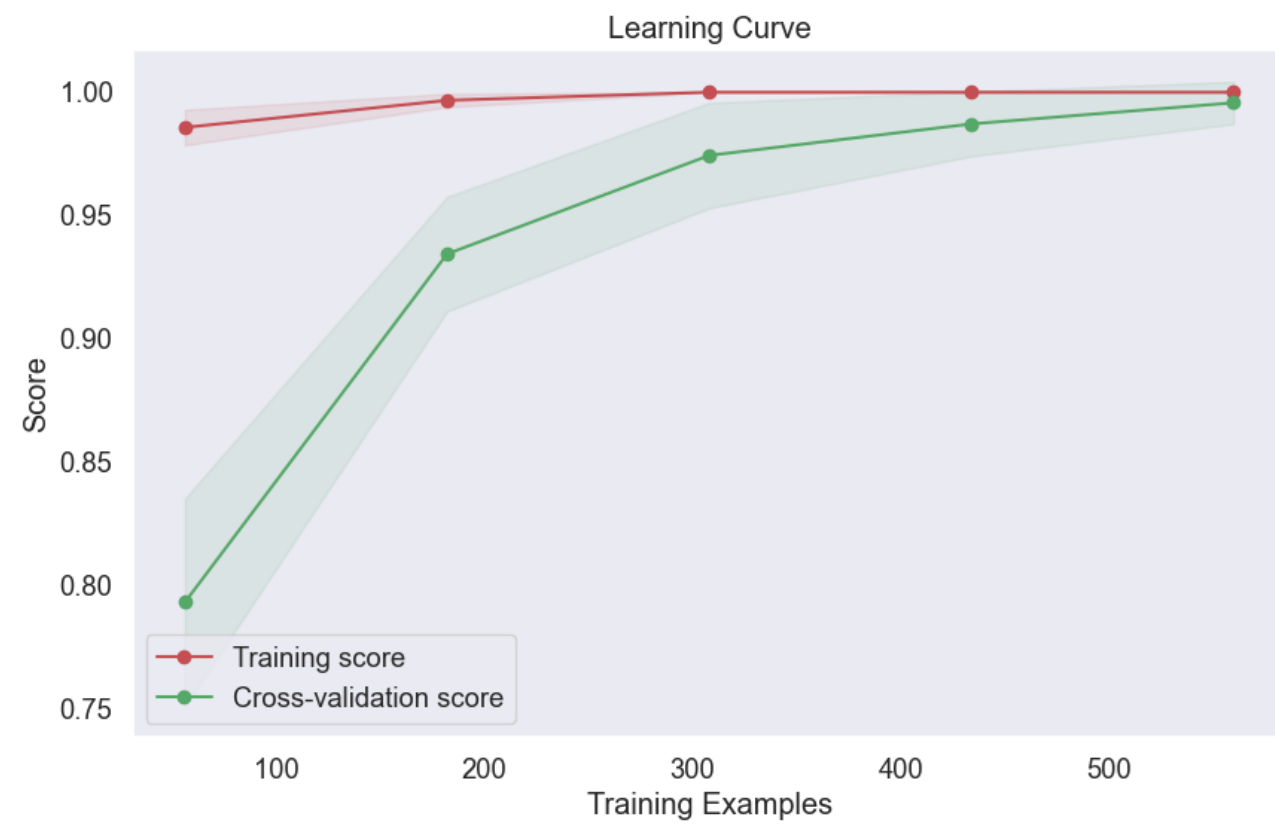

**Figure 23.** Learning Curve SVM with RBF Kernel.

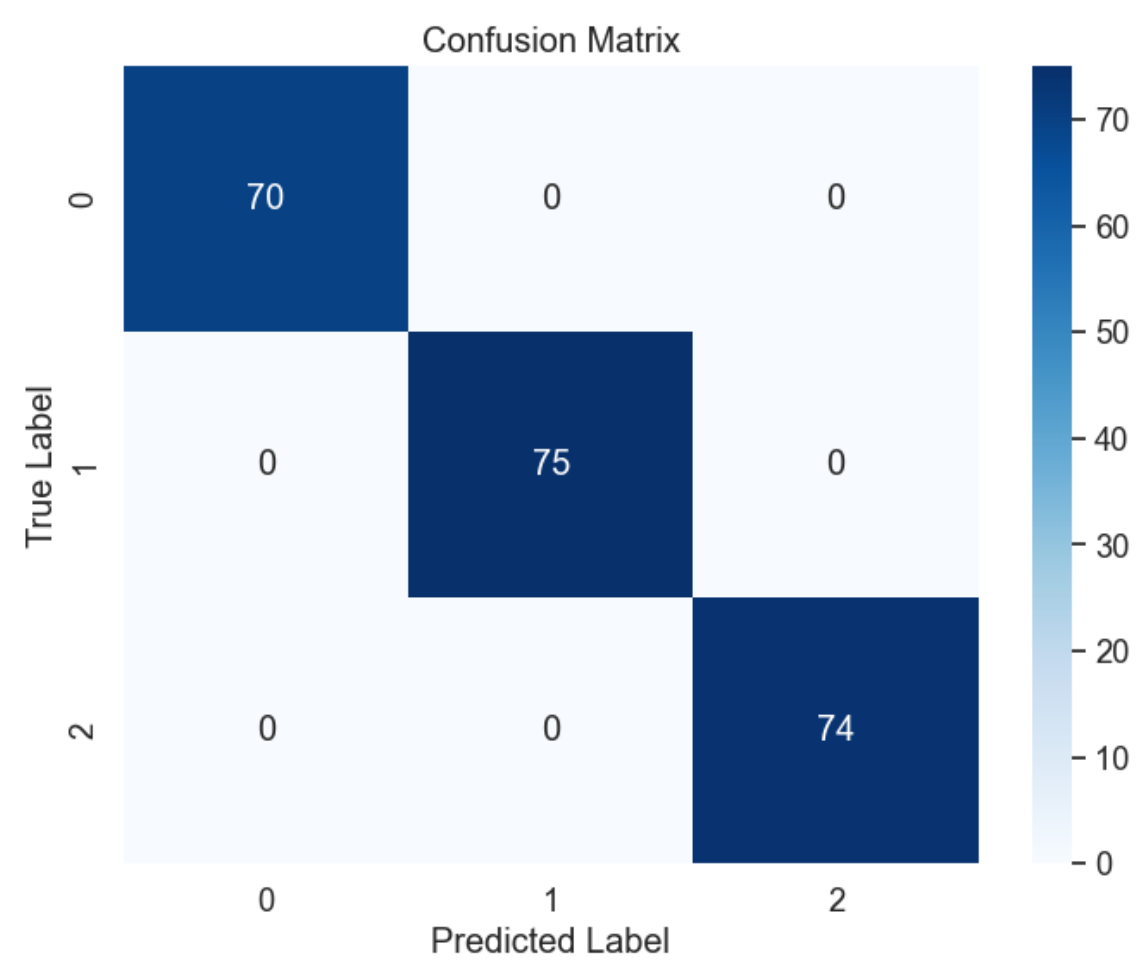

**Figure 24.** Confusion Matrix of SVM with Linear Kernel.

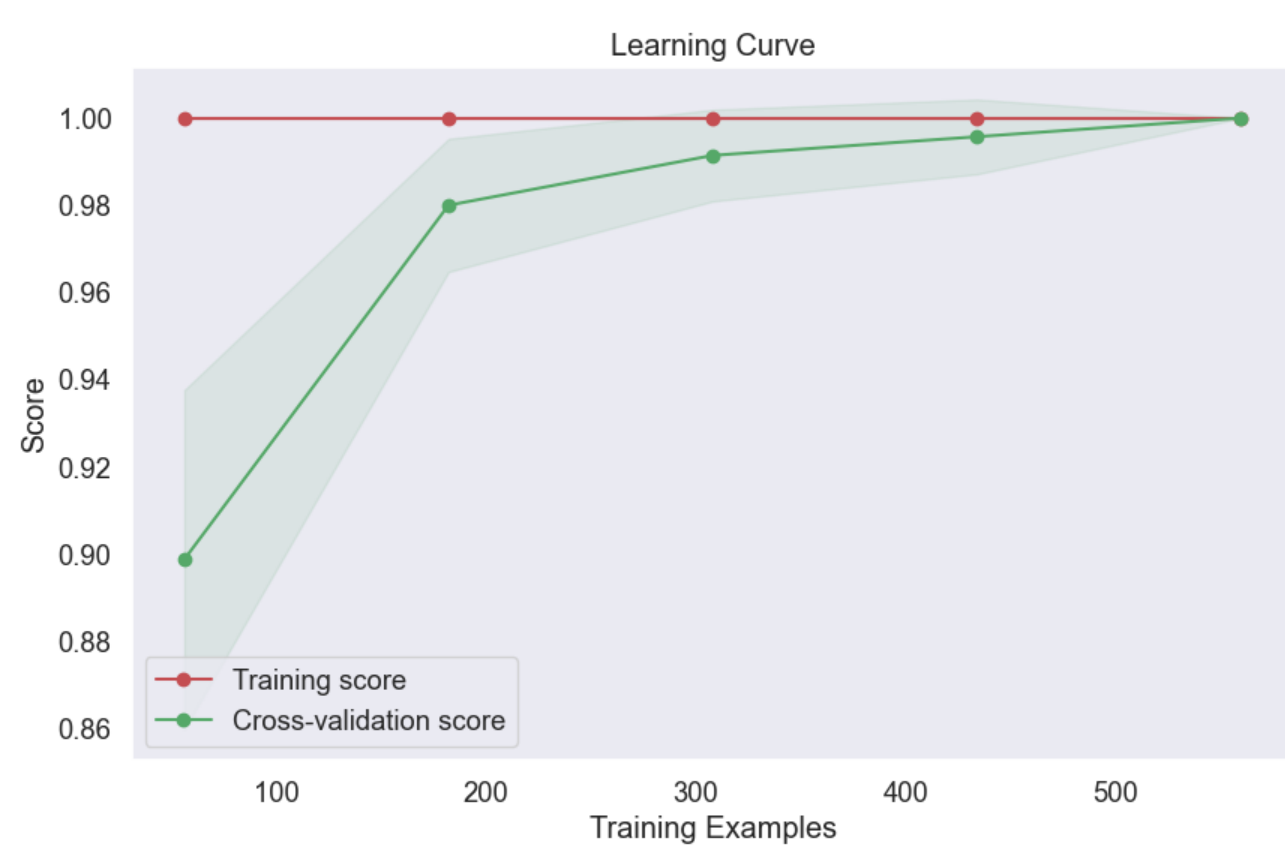

**Figure 25.** Learning Curve of SVM with Linear Kernel.

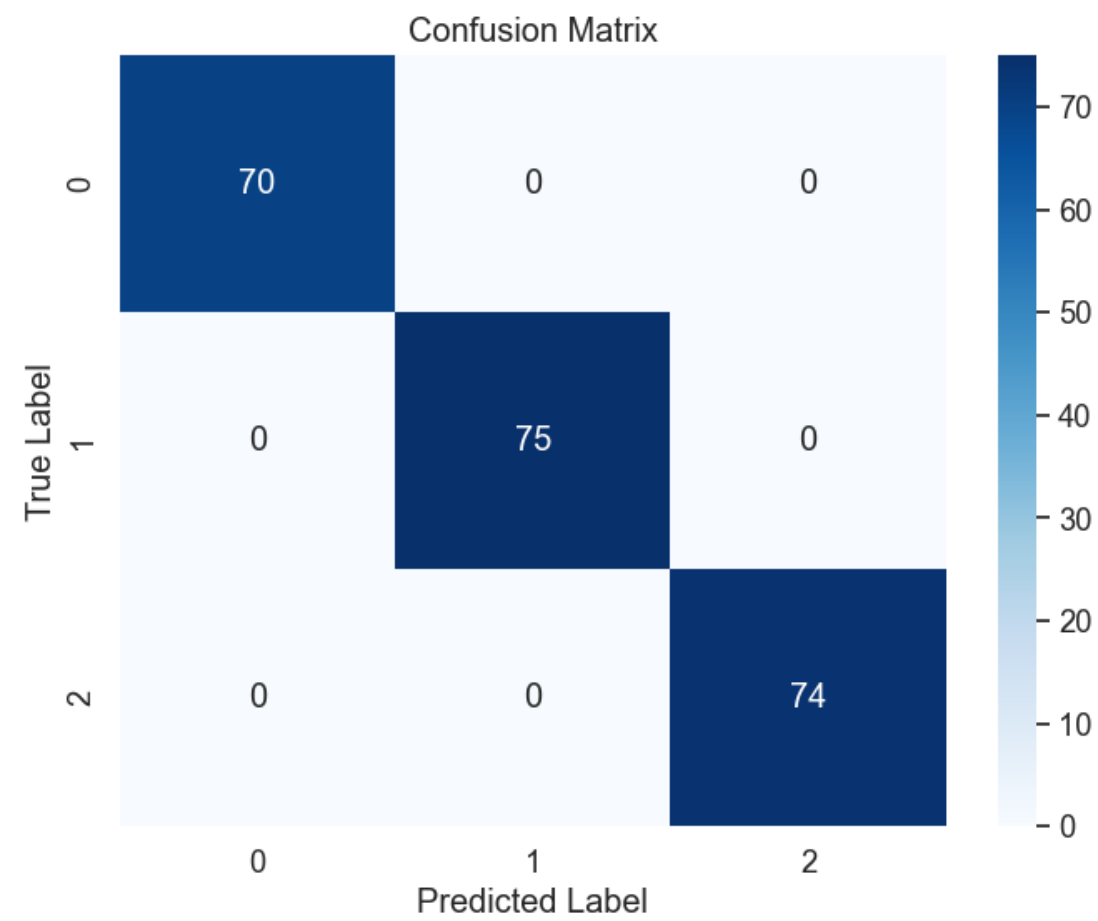

**Figure 26.** Confusion Matrix of SVM with Polynomial Kernel.

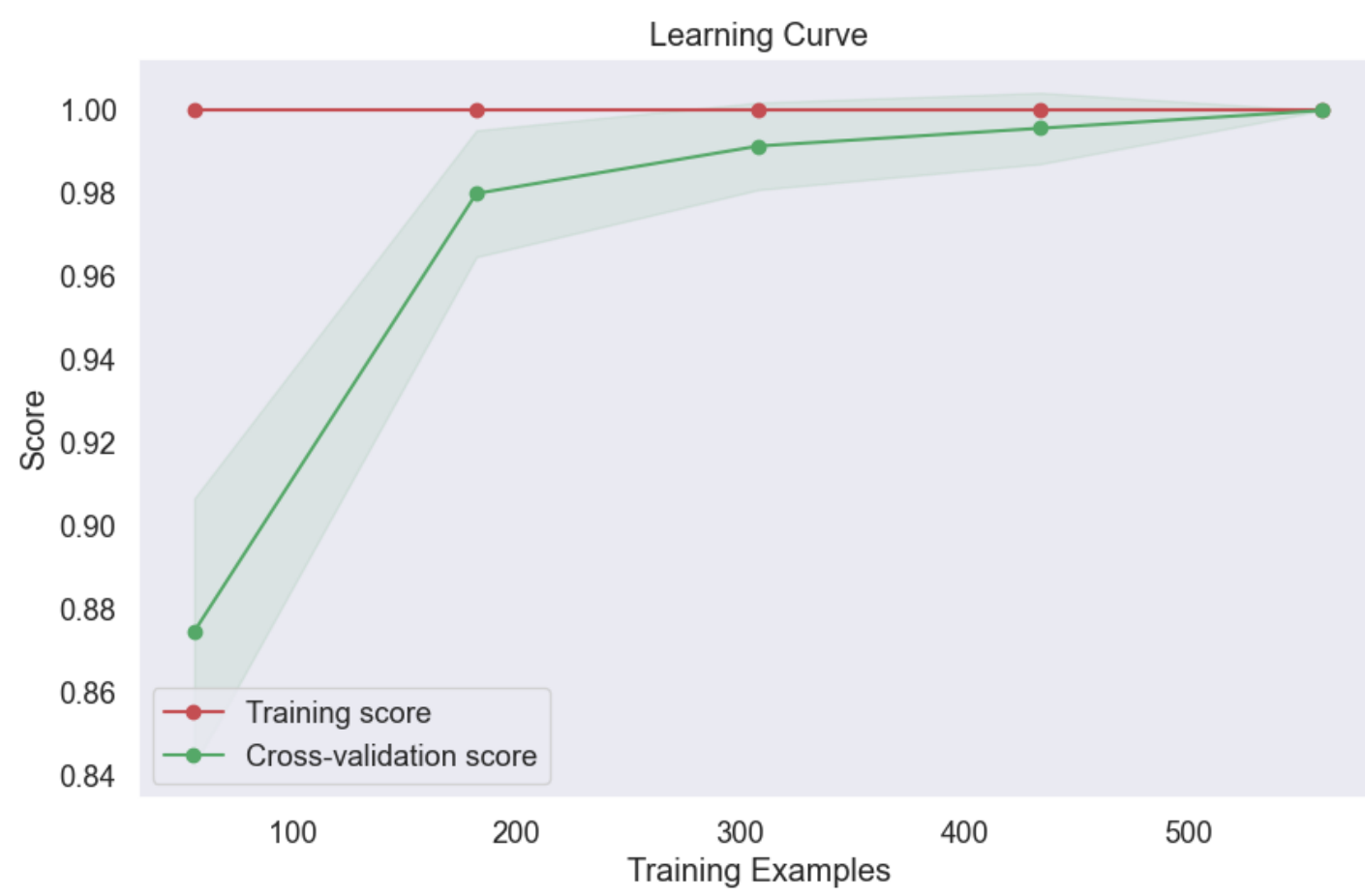

**Figure 27.** Learning Curve of SVM with Polynomial Kernel.

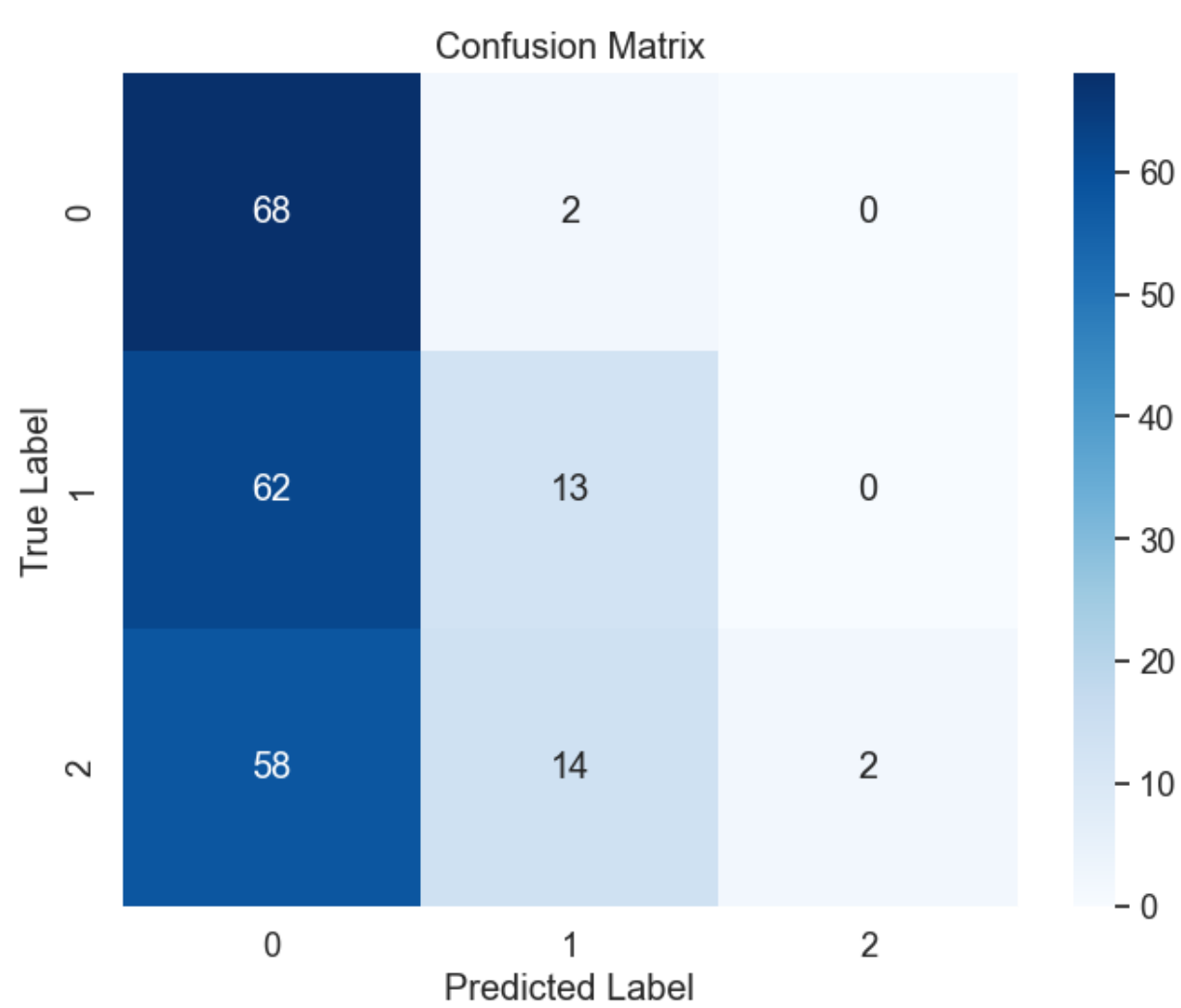

**Figure 28.** Confusion Matrix of SVM with Sigmoid Kernel.

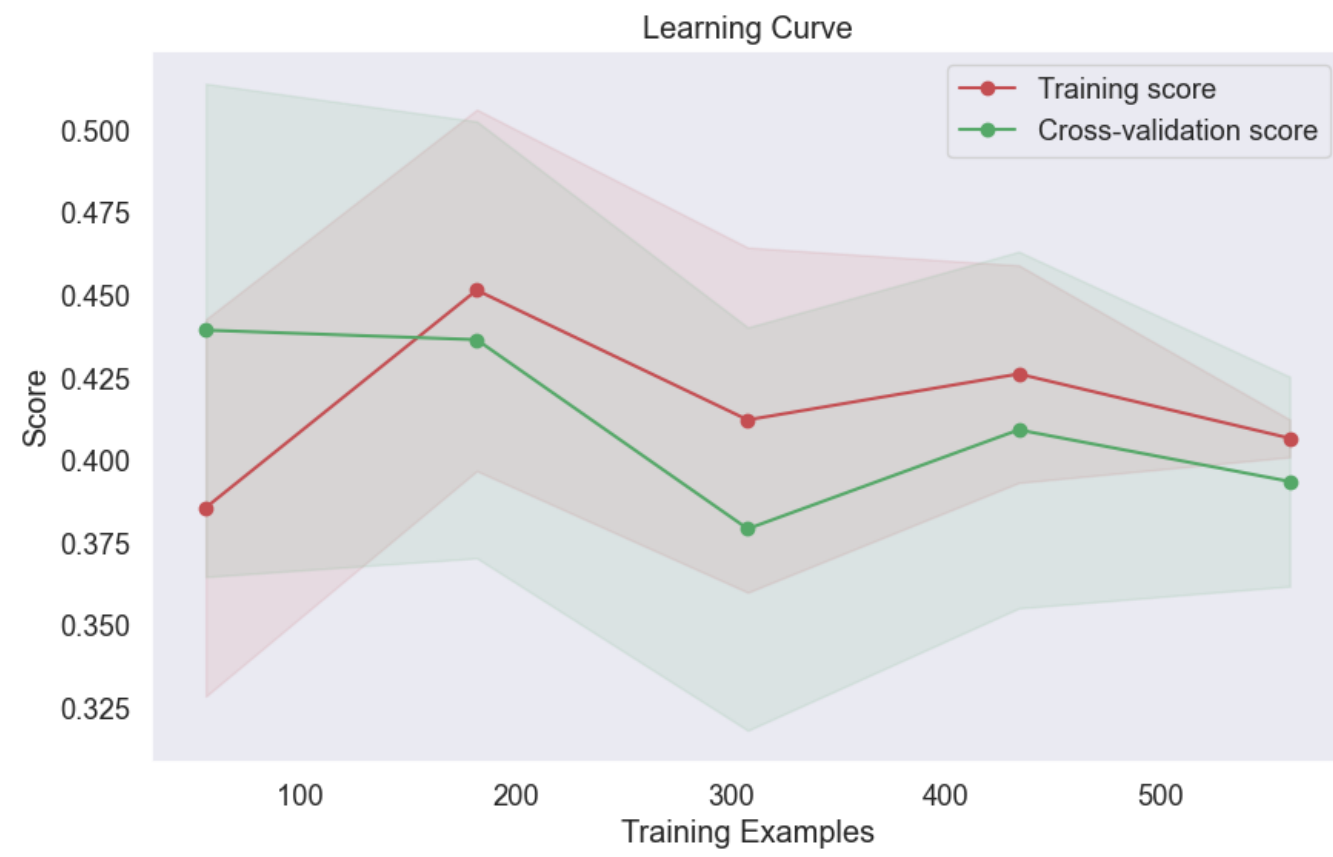

**Figure 29.** Learning Curve of SVM with Sigmoid Kernel.

In summary, our study comprehensively evaluated six machine learning (ML) models for lung cancer level classification, leveraging sophisticated monitoring techniques such as adjusting minimum child weight and learning rate to optimize model performance. ***Figure 30*** and ***Figure 31*** serve as pivotal tools for comparative analysis, presenting a comprehensive overview of the models' performance across various evaluation metrics. Through meticulous evaluation, we gained valuable insights into the strengths and weaknesses of each ML model, shedding light on their efficacy in accurately classifying lung cancer levels. These insights are crucial for informing clinical decision-making and advancing oncological diagnostics.

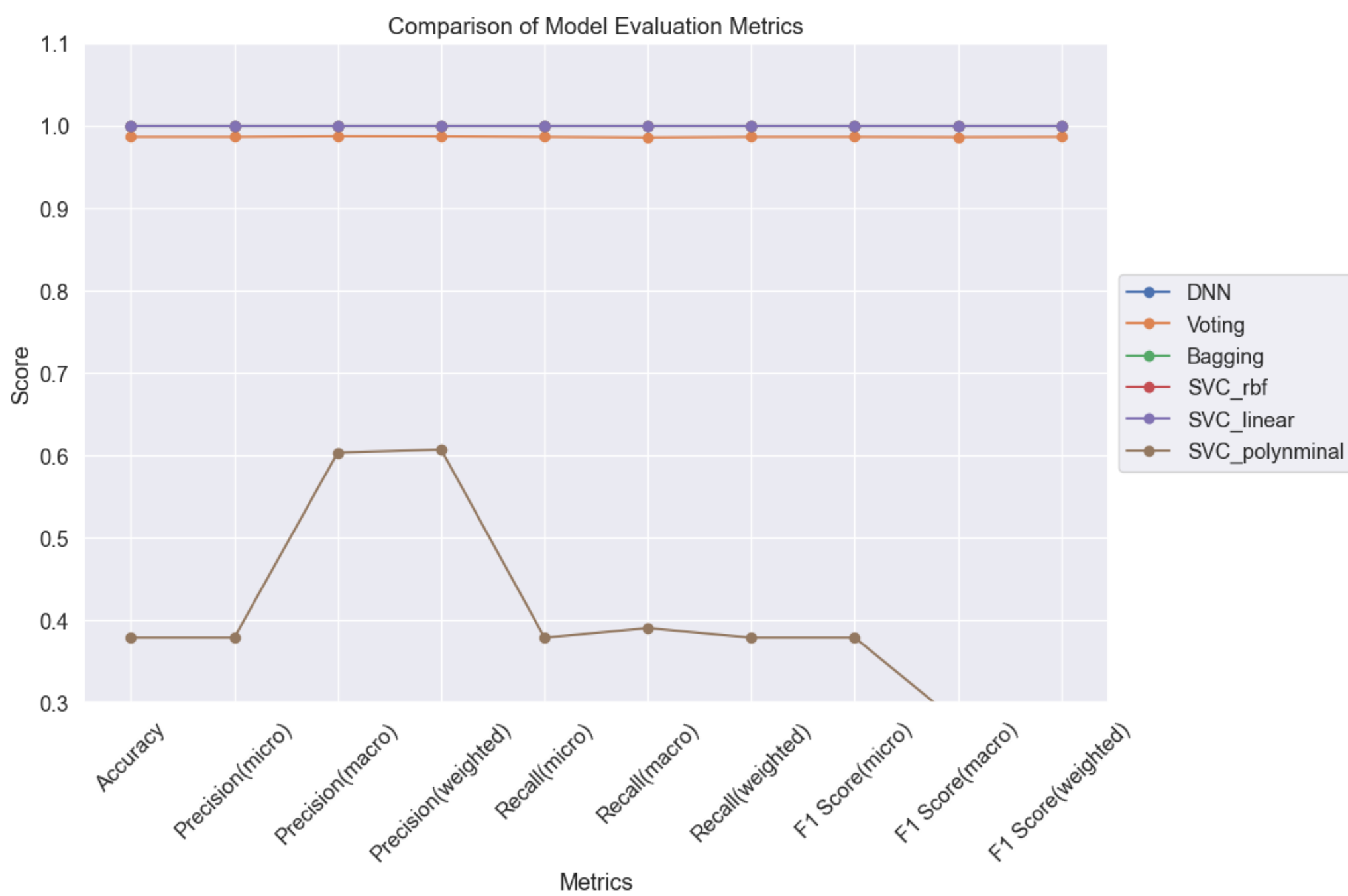

**Figure 30.** Comparison of ML Models for Lung Cancer Prediction.

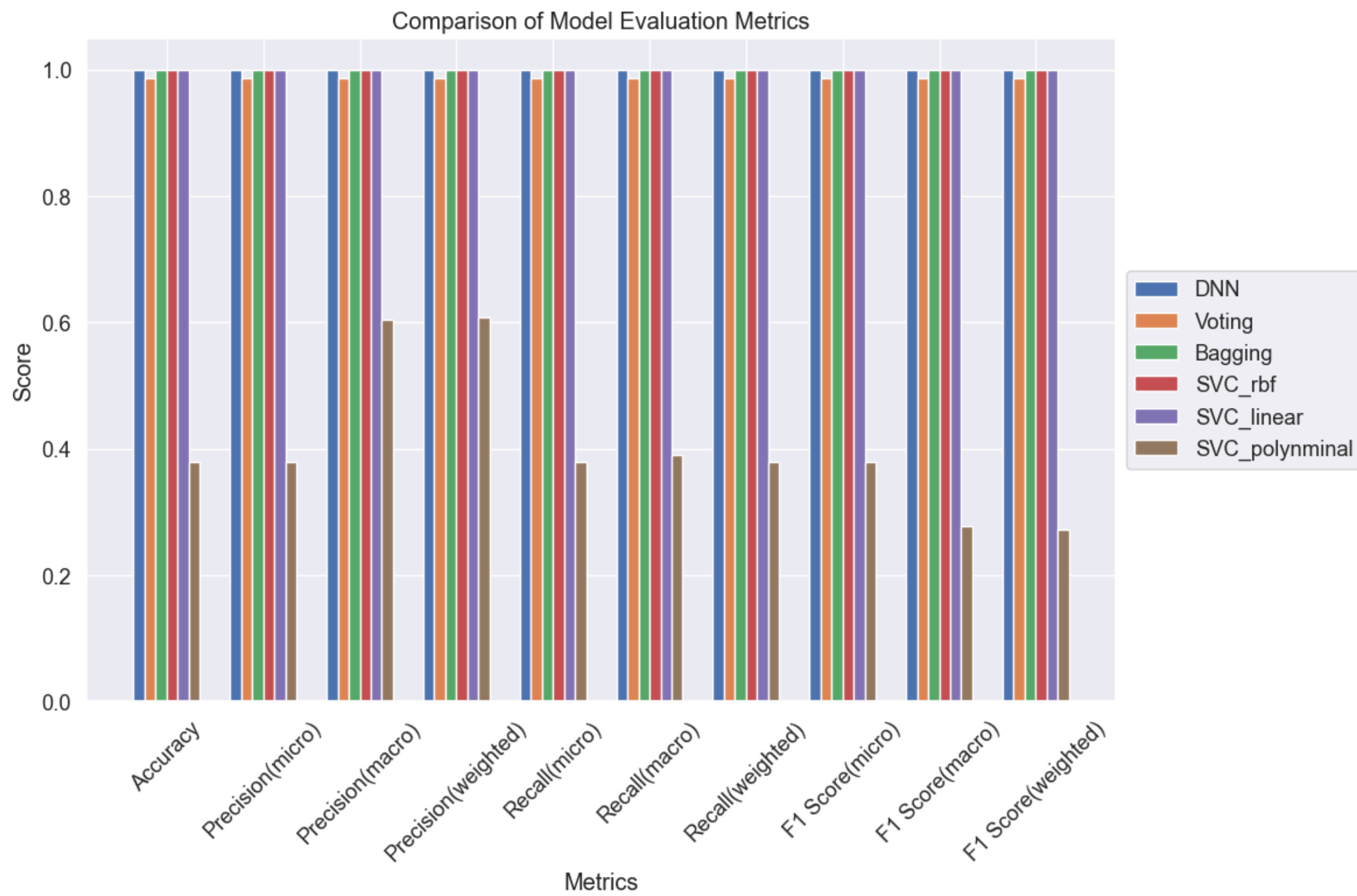

**Figure 31.** Comparison of 6 ML Models for Lung Cancer Prediction.

Moving forward, our findings provide a solid foundation for further research aimed at refining ML algorithms and techniques for lung cancer classification. By harnessing the power of advanced analytics and machine learning, we can continue to enhance the accuracy and reliability of lung cancer diagnosis, ultimately improving patient outcomes and advancing the field of oncology.

# Conclusion

In conclusion, our journey through the intricate landscape of lung cancer level classification using machine learning has been both illuminating and promising. We've navigated through a myriad of models and techniques, uncovering hidden patterns and revealing novel insights along the way. At the forefront of our discoveries stands the formidable Deep Neural Network (DNN), a beacon of hope in the realm of oncological diagnostics. Its remarkable ability to decipher complex data patterns underscores the transformative potential of artificial intelligence in revolutionizing lung cancer diagnosis. But our quest didn't stop there. We delved deeper, exploring the synergistic power of ensemble methods like voting and bagging, where the collective wisdom of multiple models converges to achieve unparalleled accuracy and reliability. Yet, amidst our triumphs, we encountered challenges. The enigmatic Support Vector Machine (SVM) with the Sigmoid kernel posed a formidable adversary, reminding us of the importance of humility and adaptability in the face of complexity. Through it all, our journey has been guided by the steady hand of methodological rigor. From fine-tuning hyperparameters to meticulous model monitoring, we've remained steadfast in our pursuit of excellence, ensuring that every step taken brings us closer to our ultimate goal: improving patient outcomes and advancing the frontier of oncological care. As we bid farewell to this chapter of our research, we do so with optimism and resolve. For in the crucible of scientific inquiry lies the promise of a brighter future, where the tools forged today pave the path to tomorrow's breakthroughs. So let us embark on this journey together, armed with knowledge, curiosity, and the unwavering belief that through innovation and collaboration, we can overcome even the greatest of challenges. The road ahead may be long and arduous, but with determination as our compass, we march forward, ever closer to a world where lung cancer is not just treatable, but preventable.

**Conflict of interest**

The authors declare no conflict of interest